\pdfoutput=1

\documentclass[11pt]{article}

\usepackage[]{acl}

\usepackage{times}
\usepackage{latexsym}

\usepackage[T1]{fontenc}

\usepackage[utf8]{inputenc}

\usepackage{microtype}

\usepackage{inconsolata}

\usepackage{amsmath,amsfonts,bm}
\usepackage{graphicx}
\usepackage{multirow}
\usepackage{makecell}
\usepackage{array}
\usepackage[para]{threeparttable}
\usepackage{booktabs}
\usepackage{tabularx}
\usepackage{xspace}
\usepackage{caption}
\usepackage{subcaption}
\usepackage{longtable}
\usepackage{hyperref}
\usepackage{bbm}
\usepackage{bm}
\usepackage{enumitem}
\usepackage{algorithm}
\usepackage{algpseudocode}
\usepackage{geometry}
\usepackage{comment}
\usepackage[colorinlistoftodos]{todonotes}
\usepackage{tablefootnote}
\usepackage{color}
\usepackage{tcolorbox}
\usepackage{tikz}
\usepackage{tabularray}
\usepackage[normalem]{ulem}

\usepackage[T1]{fontenc}
\usepackage[utf8]{inputenc}
\usepackage{listings}
\usepackage{xcolor}
\usepackage{xspace}

\newcommand{\ours}{{semiparametric token-sequence co-supervision}\xspace}
\newcommand{\oursm}{\textsc{NTP + NSP}\xspace}
\newcommand{\ret}{\textsc{[nsp]}\xspace}
\newcommand{\cs}{\textsc{[Cs]}\xspace}
\newcommand{\ce}{\textsc{[Ce]}\xspace}
\newcommand{\tool}{\textsc{NTP}\xspace}
\newcommand{\ctx}{$\texttt{Emb}_{seq}$\xspace}
\newcommand{\gen}{\texttt{Gen}\xspace}
\newcommand{\ntp}{$L_{\text{NTP}}$\xspace}
\newcommand{\nsp}{$L_{\text{NSP}}$\xspace}

\definecolor{main}{HTML}{5989cf}    
\definecolor{sub}{HTML}{cde4ff}     

\tcbset{
    sharp corners,
    colback = white,
    before skip = 0.2cm,    
    after skip = 0.5cm      
}    

\newtcolorbox{boxC}{
    colback = sub, 
    boxrule = 0pt  
}

\title{Semiparametric Token-Sequence Co-Supervision}

\author{Hyunji Lee$^1$$^{\dag}$ \quad Doyoung Kim$^1$$^{\dag}$ \quad Jihoon Jun$^4$\thanks{\hspace{1mm}Work performed during internship at KAIST.} \quad Sejune Joo$^1$ \quad  \\ \textbf{Joel Jang$^2$} \quad \textbf{Kyoung-Woon On$^3$} \quad \textbf{Minjoon Seo$^1$} \\
 \\
$^1$KAIST AI \quad $^2$University of Washington \quad $^3$Kakao Brain \quad $^4$ Seoul National University \\
\texttt{\{hyunji.amy.lee, doyoungkim, minjoon\}@kaist.ac.kr} 
}

\begin{document}
\maketitle
\def\thefootnote{$^{\dag}$}\footnotetext{Denotes equal contribution}
\def\thefootnote{\arabic{footnote}}
\begin{abstract}

In this work, we introduce a semiparametric token-sequence co-supervision training method. It trains a language model by simultaneously leveraging supervision from the traditional next token prediction loss which is calculated over the parametric token embedding space and the next sequence prediction loss which is calculated over the nonparametric sequence embedding space. The nonparametric sequence embedding space is constructed by a separate language model tasked to condense an input text into a single representative embedding. 
Our experiments demonstrate that a model trained via both supervisions consistently surpasses models trained via each supervision independently. 
Analysis suggests that this co-supervision encourages a broader generalization capability across the model. 
Especially, the robustness of parametric token space which is established during the pretraining step tends to effectively enhance the stability of nonparametric sequence embedding space, a new space established by another language model\footnote{Code and Data in \url{https://github.com/kaistAI/Semiparametric_Token-Sequence_Co-Supervision}}.
\end{abstract}

\section{Introduction}

Language models are typically trained through next-token prediction (NTP), where the model forecasts the distribution of the next token's embedding, given a current token embedding~\cite{touvron2023llama, brown2020language, zhang2022opt}. This process relies on a language model head, which includes embeddings for the entire vocabulary. While this approach has demonstrated high performance, its reliance on predicting over a finite parametric vocabulary space restricts the models' expressivity~\citep{min2022nonparametric, yang2017breaking, pappas2020grounded}. 
Also, such supervision constrains the model's predictive capabilities to only the next token, whereas humans can anticipate sequences of varying granularities highlighting a significant gap.

In this work, we aim to enhance the capabilities of language models by superposing parametric token embedding space and nonparametric sequence embedding space at the output space of a language model. Drawing on previous research that highlights the adaptable nature of language models' parametric token embedding space~\citep{li2021prefix, lester2021power, Hao2023ToolkenGPTAF}, we theorize that the language model can integrate a new embedding space alongside the model's existing parametric space and can also leverage the stable foundation of its parametric space established during the pretraining phase when integrating the new embedding space.

\begin{figure}[t!]
    \centering
    \begin{minipage}[b]{0.45\textwidth}
    \includegraphics[width=\textwidth]{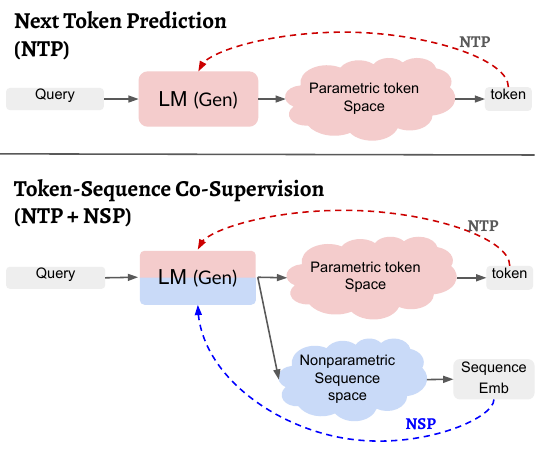}
    \end{minipage}
\caption{\fontsize{6.5}{10}\footnotesize 
While previous methods train language models with next token prediction loss (NTP), \ours trains a language model in a \textit{multi-task} manner where supervision from parametric token embedding space (NTP) and supervision from nonparametric sequence embedding space (NSP) flow simultaneously.
}
\label{fig:overall}
\end{figure}

To this end, we introduce \ours, a novel training approach that trains a language model (\gen) by incorporating supervision from \textit{both} the traditional next token prediction (NTP), calculated over the parametric token embedding space, and next sequence prediction (NSP), which is calculated over nonparametric sequence embedding space as in Figure~\ref{fig:overall}.
This nonparametric sequence embedding space is constructed by another language model, \ctx, which compresses the entire input text into a singular, representative embedding. The supervision is calculated via contrastive learning over embedding from the nonparametric embedding space and the output distribution from \gen.

We experiment across 10 information-seeking datasets from the KILT~\citep{petroni-etal-2021-kilt} and ALCE~\citep{gao2023enabling} benchmarks. We compare a model trained via \ours, which employs multi-task training over NTP and NSP, against a model trained under each supervision individually. 
The findings reveal that models under co-supervision significantly outperform those trained with separate supervision, with an average performance improvement of 14.2. 
This demonstrates that constructing a common space through co-supervision fosters the generalization and robustness of the language model. 
The nonparametric space under \ours is more stable compared to models trained solely on NSP, suggesting that the robustness of the parametric space, established through pretraining, provides a solid foundation that enhances the stability of the nonparametric space. Also, unlike models trained only via NTP, the model trained via \ours tends to effectively use knowledge from the nonparametric space during generation, suggesting a shift from rote learning to active knowledge utilization. 

Notably, models trained via token-sequence co-supervision exhibit significant improvements, particularly in out-of-domain datasets, with an average enhancement rate of 6.6 over in-domain datasets. 
Also, this co-supervision fosters a strong interaction between the parametric token and nonparametric sequence spaces, with \gen more inclined to generate responses by drawing upon knowledge from the nonparametric space. 
Moreover, the inherent distribution of \ctx established during a pretraining step influences the overall performance where aligning \ctx and \gen to the same pretrained LM thereby the same distribution contributes to a more stable training process.

\section{Related Works}

\paragraph{Aligning two different models}

Various studies have explored ways to align two different models. 
In multi-modal tasks, efforts have been made to connect pretrained vision-only and language-only models, either by employing cross-attention mechanisms~\citep{alayrac2022flamingo} or by aligning the vision encoder's output embedding with the language model's input space~\citep{liu2023llava}.
Also, aligning a language model with another multilingual language model enables the model to perform multilingual tasks by leveraging the distribution from the multilingual model~\citep{bansal2024llm, yoon2024langbridge}. 
Moreover, recent research has focused on retrieval-augmented language models that align a retrieval model with a language model to mitigate the issue of hallucination in language models. 
Studies suggest that aligning these two models leads to improved performance compared to training them separately~\citep{Lin2023RADITRD, shi2023replug}.
\ours also aims to train on aligning two different models but is unique in that it focuses on aligning the two models at the output space of the language model rather than the input space. 

\begin{figure*}[t!]
    \centering
    \begin{minipage}[b]{0.9\textwidth}
    \includegraphics[width=\textwidth]{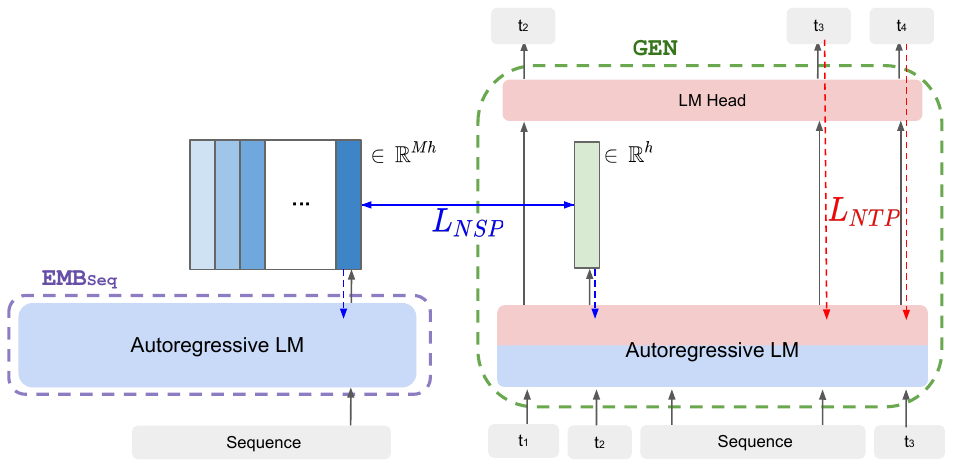}
    \end{minipage}
\caption{\fontsize{6.5}{10}\footnotesize Overview of semiparametric token-sequence co-supervision. \gen is an autoregressive LM with LM head on top, which is trained with co-supervision over parametric token embedding space (\ntp) and nonparametric sequence embedding space (\nsp). \ctx, another autoregressive LM constructs nonparametric sequence embedding space with the output embeddings when given sequence as input. $t_i$ indicates tokens, $h$ indicates dimension size of hidden state, and $M$ indicates number of sequences in a batch (Refer Appendix~\ref{app: train} for datailed calculation).} 
\label{fig:cosup}
\end{figure*}

\paragraph{Language Models with Nonparametric Embeddings}

Integrating nonparametric embeddings into language models has consistently demonstrated advantages. 
This approach enhances the expressiveness beyond the inherent capabilities of language models~\citep{khandelwal2019generalization, zhong2022training}. 
Also, leveraging the rich contextual knowledge through nonparametric embeddings effectively reduces instances of hallucination and improves the generation of accurate and factual content~\citep{Lewis2020RetrievalAugmentedGF, borgeaud2022improving, guu2020retrieval}. 
Moreover, it shows high performance for rare and unseen cases as such tend to not exist in model parametric space~\citep{lee2023nonparametric, min2022nonparametric}.
\ours also leverages the nonparametric embedding space but is unique in that it \textit{trains} the model to utilize \textit{both} the parametric and nonparametric embedding space, where the nonparametric embedding space is not static at the training step, but is trainable making the nonparametric space more adaptable to the well-constructed parametric embedding space.

\section{Semiparametric Token-Sequence Co-Supervision}

In Section~\ref{sec3: NTP}, we delve into our interpretation of the Next Token Prediction (NTP), first laying the foundation for our hypothesis. Subsequently, in Section~\ref{sec3: other}, we explore next sequence prediction (NSP), an extension of next token prediction to nonparametric sequence embedding space. In Section~\ref{sec3: semiparasup}, to test our hypothesis we introduce Semiparametric token-sequence co-supervision, a training method with supervision from both parametric token embedding space (NTP) and nonparametric sequence embedding space (NSP) in a simultaneous manner.

\subsection{Revisiting Next Token Prediction} \label{sec3: NTP}

We revisit the conventional approach of Next Token Prediction (NTP), which forms the foundation of most modern language models (LMs). 
NTP is a process to predict token $t$ over the vocabulary set $V$ when given the preceding tokens $t_1, \ldots, t_k$ to a language model:

\begin{align}
\text{argmax}_{t\in V} \ P(t | t_1, \ldots, t_k)
\label{eq:NTP}
\end{align}

In more detail, as shown in Figure~\ref{fig:cosup}, a language model (\gen) consists of a language model (Autoregressive LM) and language model head (LM Head). 
LM Head $\bold{W}_V$ ($\in \mathbb{R}^{|V| \times h}$) is a linear layer where $h$ denotes the model hidden state dimension.
When given sequence of tokens $t_1, \ldots, t_k$ to LM, it returns a hidden state vector $\textbf{q}_k$ ($\in \mathbb{R}^h$). The hidden state is calculated with LM Head which returns the probability distribution over vocabulary size. Thereby Equation~\ref{eq:NTP} can be reformulated as:

\begin{align}
\text{argmax}_{t\in V} \bold{W}_V \textbf{q}_k
\label{eq: NTP2} 
\end{align}

Equation~\ref{eq: NTP2} can be interpreted as a retrieving stage, indicating that the parametric token embedding space $\textbf{W}_V$ (LM Head) which consists of model vocab set determines the corpus (the range of objects from which "what" we will retrieve) when given $\textbf{q}_k$, the hidden state embedding as a query. 
As the conventional language modeling paradigm only provides supervision over a fixed vocabulary-sized token embedding space $\textbf{W}_V$, the usage was limited to predicting the most probable next token embedding. 
However, with such interpretation, when given multiple supervision from various embedding spaces, the methodology is extendable to predicting not only token embedding $\textbf{W}_V$ but also various representatives in any other non-parametric embedding spaces.

\subsection{Next Sequence Prediction} \label{sec3: other}

Broadening the scope of next token prediction (NTP), we explore the domain of sequence-level embedding space $\textbf{W}_\mathcal{C}$. 
In natural language processing, many tasks extend beyond merely predicting the next token; they necessitate the utilization of sequence-level knowledge such as contexts from external memory from a corpus~\citep{zhong2022training, Hao2023ToolkenGPTAF, Lewis2020RetrievalAugmentedGF}. This is where the next sequence prediction (NSP) becomes invaluable. Unlike traditional NTP, which operates over parametric token embedding space, NSP interacts with nonparametric sequence embedding space. 
NSP allows models to anticipate and generate answers based on given \textit{sequences} on-demand, mirroring the human-like ability to anticipate sequences of varying granularities and its ability to refer to external sequences while answering a question.

When equipped with an embedding space $\textbf{W}_\mathcal{C}$ representing embedding space constructed with a corpus set of sequences $\mathcal{C}$, NSP is a process to predict the sequence $s$ akin to Equation~\ref{eq: NTP2}.
\begin{align*}
\text{argmax}_{s\in \mathcal{C}} \bold{W}_{\mathcal{C}} \textbf{q}_k 
\end{align*}

\noindent Specifically, as shown in Figure~\ref{fig:cosup}, the embedding space $\textbf{W}_\mathcal{C}$ is constructed by the last hidden state embedding of another autoregressive model \ctx; $\bold{W}_{\mathcal{C}}$ is a nonparametric embedding space constructed with a set of last hidden state embedding $\textbf{s}$ where $\textbf{s} = \text{\ctx}(s)$ for $ s\in \mathcal{C}$. 

There are previous studies that explore supervision beyond the token level, targeting higher levels of text granularity~\citep{devlin2018bert, joshi2020spanbert}. However, their motivations are distinct from \ours; our objective of additional supervision (NSP) is to leverage co-supervision to foster a unified space that bridges the nonparametric sequence embedding space with the parametric token embedding space. In contrast, earlier efforts, such as the next sentence prediction (NSP) feature in BERT~\citep{devlin2018bert}, focused on a simpler binary supervision task that determines the relevance of a succeeding sentence. Also to the best of our knowledge, \ours is the first approach to train autoregressive language model with co-supervision apart from only NTP.

\subsection{Co-Supervision} \label{sec3: semiparasup}

We introduce semiparametric token-sequence co-supervision, a training approach that trains a language model (\gen) by incorporating supervision from both the traditional next token prediction (NTP) which is calculated over the parametric token embedding space, and next sequence prediction (NSP) which is calculated over nonparametric sequence embedding space as shown in Figure~\ref{fig:cosup}. 

For NTP, we apply an ordinary casual language modeling loss function. When given input $X$:
\begin{align}
L_{\text{NTP}} = -\frac{1}{|X|} \sum_{t_i\in X}^{}{log P_{\text{\gen}}(t_i|t_{<i})}
\label{eq: ntp_loss}
\end{align}

For NSP, we apply the contrastive InfoNCE loss~\citep{karpukhin2020dense, izacard2021contriever}; when given a query embedding, positive sequence pair relevant to the query, and a pool of negative sequences unrelated to the query, \ctx and \gen are trained to maximize the similarity between the query embedding and the positive pair and minimize the similarity with the negatives pairs where the query embedding is last hidden state embedding of \gen and sequence embeddings are set of embeddings from the last hidden state of \ctx. 
As it is impractical to dump all embeddings of a corpus set every step, here we approximate softmax over all corpus by softmax over positive and negative pairs in the same batch~\citep{karpukhin2020dense, izacard2021contriever}; we consider other sequences in the same batch as negatives (in-batch negatives). 
Formally, given a query embedding $\textbf{q}$, positive pair passage embedding $\textbf{c}_i^+$, and negative pairs ${\textbf{c}_{1}^-, \cdots, \textbf{c}_{M-1}^-}$ where $M$ is the number of sequences in a batch, NSP is calculated via:
\begin{align}
L_{\text{NSP}}(\textbf{q}_i, \textbf{c}_i^+, \textbf{c}_{1}^-, \ldots, \textbf{c}_{M-1}^-) \notag\\ = -\log \frac{e^{\text{sim}(\textbf{q}_i,\textbf{c}_i^+)}}{e^{\text{sim}(\textbf{q}_i,\textbf{c}_i^+)} + \sum_{j=1}^{M-1} e^{\text{sim}(\textbf{q}_i,\textbf{c}_{j}^-)}}
\label{eq: nsp_loss}
\end{align} 

Thereby total loss over semiparametric token-sequence co-supervision is calculated as: 
\begin{align}
L_{\text{co-supervision}} = \text{\ntp} + \lambda \text{\nsp}
\label{eq: cosup_loss}
\end{align}

\noindent where $\lambda$ is the weight parameter to match the loss scale between \nsp and \ntp as it flows through \gen together.

\section{Implementation Details} \label{semicosup}
In this section, we share implementation details of experiments over information-seeking datasets.
In Section~\ref{sec3: notations}, we share details of the problem setup.
In Section~\ref{sec3: training}, we describe details of how we train a language model (\gen) with both supervision of next token prediction (NTP) and next sequence prediction (NSP) simultaneously, and the inference step of the trained model in Section~\ref{sec3: inference}. More details are in Appendix~\ref{app: implementation_details}.

\subsection{Problem Setup} \label{sec3: notations}

For details, we start with an explanation of notations and training instances.
\gen is the trainable language model that trains over by co-supervision from both NTP and NSP. 
\ctx is the trainable language model that constructs the nonparametric sequence embedding space and which calculates over \gen output embedding for NSP loss.

Training instances are in a form of ``$q, n_1, \ret, c_1, n_2, \cdots n_i, \ret, c_i, \cdots $''~\citep{asai2023selfrag} (Examples in Appendix~\ref{app: training_dataset}). 
$q$ is an input query. 
\ret is a special token indicating the model to calculate over sequence embedding space, in other words when the model itself considers external knowledge is necessary. 
$c_i$ is a relevant (positive) sequence when given sequences before \ret as input. For example, $c_1$ is the relevant sequence when given $q, n_1$ as an input.
The number of \ret varies from 0 to multiple differing by how many times the query necessitates external knowledge during generation.

\subsection{Training} \label{sec3: training}

Figure~\ref{fig:cosup} shows the overview of how we train a language model \gen with \ours, where both \ntp~(equation~\ref{eq: ntp_loss}) and \nsp~(equation~\ref{eq: nsp_loss}) flows through a language model \gen simultaneously. 
For \nsp loss, we train another language model \ctx together which constructs a nonparametric sequence embedding space. 
Specifically, the query embedding ($\textbf{q}$) and sequence embeddings ($\textbf{c}$) to calculate \nsp are calculated by:

\begin{align}
\textbf{q} &= \text{\gen}\left(\text{query} \ret\right)[-1] \label{eq:q} \\
\textbf{c} &= \text{\ctx}\left(\text{<s>} \text{sequence} \text{</s>}\right)[-1] \label{eq:c}
\end{align}
\noindent which is the last layer token representation of \ret from \gen and the last layer token representation of end-of-sequence token (</s>) from \ctx, respectively.
Thereby the gradient of \nsp flows through both \ctx and \gen. 
Whereas for \ntp, the gradient only flows through \gen.

\subsection{Inference} \label{sec3: inference}

During the inference step, we first dump all sequence embeddings with \ctx during the offline time.
Given a set of sequences of size $M$, we feed each sequence into \ctx and extract representative embedding $\textbf{c}$ from which we get a context embedding matrix of $\textbf{C} \in \mathbb{R}^{h \times M}$. 

After dumping, \ctx is no longer necessary and we only need \gen.
The generated response is in the same form as the training instances ``$q, n_1, \ret, c_1, n_2, \cdots n_i, \ret, c_i, \cdots $'' (Section~\ref{sec3: notations}). 
When \gen generates \ret, it treats the last layer representation of \ret, generated after inputting sequences up to \ret ($q, \cdots, n_i$), as the query embedding $\textbf{q}$. This embedding $\textbf{q}$ is then calculated over the context embedding matrix $\textbf{C}$ to find the most relevant sequence ($c_i$) for the query. 
$c_i$ is added after \ret, allowing the generation process to proceed based on the sequence. When \gen generates a token other than \ret, it functions the same as a standard language model, selecting the next token based on the highest probability through the language modeling head.

\section{Experiments Setup} \label{exp_setup}
In this section, we share the experimental setup of baseline, metric, training and evaluation dataset, and training details.
More details of each paragraph are in Appendix~\ref{sec:exp_setup}.

\paragraph{Baseline}

For the analysis of how co-supervision affects model performance, we train a baseline model using the same approach as outlined in Section~\ref{sec3: training}, but with each supervision (NTP and NSP) \textit{separately}. NTP is computed in the same manner for both \ours and the baseline, involving the flow of gradients through \gen. However, NSP is calculated differently: while \ours computes it between the output embeddings of \gen (Eq~\ref{eq:q}) and \ctx (Eq~\ref{eq:c}) thereby flowing the gradient through both models, the baseline computes it between the output embeddings of \ctx only, where \textbf{c} is the same and \textbf{q} is different:

\begin{align*}
\textbf{q} = \text{\ctx}\left(\text{query} \ret\right)[-1] \\
\textbf{c} = \text{\ctx}\left(\text{<s>} \text{sequence} \text{</s>}\right)[-1] \\
\end{align*} 

\noindent which limits the gradient of NSP to only \ctx.

\paragraph{Metric}
To measure how training a language model with both supervision of NTP and NSP shows different aspects over the model trained with each supervision separately, we measure model performance via three metrics.
Correctness and grounding performance measures the generation ability of the model and how well it utilizes knowledge from the external sequence. 
Retrieval performance mainly measures how well the model finds relevant sequences.
\textit{Correctness (Cor)} evaluates how well the model generates a response thereby answering the given query for each task. For instance, in the case of question answering (QA), correctness is measured by answer accuracy (Table~\ref{table: correctness_metric} in the Appendix).
\textit{Retrieval} performance measures whether the model finds relevant paragraphs to answer the given question which requires well-constructed nonparametric sequence space. 
\textit{Grounding (Gr)} performance evaluates how well the model generates based on given external knowledge. Following the approach of previous works~\citep{gao2023enabling, Lee2023HowWD}, we use TRUE~\citep{honovich2022true}, a T5-11B model finetuned on various NLI datasets, to see whether the external knowledge entails a part of the response that is generated based on the knowledge.

\paragraph{Training Dataset}
\label{app: training_dataset}
All models undergo training using the dataset provided by \citet{asai2023selfrag}, featuring a diverse range of instruction-following datasets. The dataset contains information-seeking questions paired with long-form responses, where a set of sequences are annotated within the responses (example form in Section~\ref{sec3: notations}). As the dataset includes instances beyond our scope such as cases where the matching sequence is irrelevant to the response as \citet{asai2023selfrag} aims to train model self-critique, a filtering process was applied.  This refined the dataset to 42,932 instances suitable for our research objectives.

\paragraph{Evaluation Dataset}
We conduct evaluations on ten long-form information-seeking datasets from two benchmarks KILT~\citep{petroni-etal-2021-kilt} and ALCE~\citep{gao2023enabling}. Some of these datasets overlap with the training dataset, categorized as in-domain datasets, while others are considered out-of-domain datasets. 
While the ALCE setting shows closer alignment of our training dataset, as the benchmark does not contain annotation of relevant sequences, we adapt the KILT benchmark to conform to the ALCE setting. For this reason, we mainly focus analysis on the KILT benchmark, which we can measure all three metrics.

\paragraph{Training details}
We use pretrained Llama2 7B~\citep{Touvron2023Llama2O} as an initial model for both \gen and \ctx. We use 8 Nvidia A100 for the experiments. We also set the base hyperparameter as epoch 3, batch size 8, learning rate of 2e-5 and a decayed rate gamma of 0.85 every 1 epoch, and AdamW optimizer~\citep{loshchilov2019decoupled} with no decay across all the experiments. For experimenting \ours, we set the weight $\lambda$ as 0.01 while training and apply gradient clipping to the \gen model, with a maximum norm equal to 1.

\section{Experimental Results \& Analysis}

\begin{table*}[h]
\centering
\fontsize{6.5}{10} \selectfont
    \begin{tabular}{cc|ccc|ccc|ccc|ccc}
    \toprule
    & Retriever &  Ret  & Cor & Gr &  Ret  & Cor & Gr &  Ret  & Cor & Gr  &  Ret & Cor & Gr \\
    \midrule
    & & \multicolumn{3}{c}{NQ*} & \multicolumn{3}{c}{WoW*} & \multicolumn{3}{c}{FEVER*} & \multicolumn{3}{c}{ELI5} \\
    \midrule
    \multirow{2}{*}{Top 20}
    & \tool  & 62.0 & 49.7 & 51.5 & 31.0 & 14.7 & 44.9 & 58.0 & 57.1 & 5.8 &  30.9 & \textbf{21.8} & \textbf{9.3} \\
    & \oursm & \textbf{65.1}  & \textbf{55.7} & \textbf{62.6} & \textbf{49.8}  & \textbf{15.7} & \textbf{63.7} & \textbf{77.5} & \textbf{65.2} & \textbf{28.0} & \textbf{36.3} & 21.5 & 8.7 \\
    \midrule
    \multirow{2}{*}{Top 100}
    & \tool & 35.7 & 38.1 & 43.0 & 14.7  & 13.4 & 40.0 & 28.8  & 56.7 & 5.2 &  12.7 & \textbf{21.9} & 7.0  \\
    & \oursm & \textbf{56.8} & \textbf{50.5} & \textbf{58.7}  &\textbf{36.9}& \textbf{14.8} & \textbf{61.3}  & \textbf{66.2} & \textbf{64.3} & \textbf{26.1} & \textbf{21.0} & 21.6 & \textbf{10.2} \\
    \midrule
    \midrule
    && \multicolumn{3}{c}{zsRE} & \multicolumn{3}{c}{T-REx} & \multicolumn{3}{c}{TriviaQA} & \multicolumn{3}{c}{HotpotQA} \\
    \midrule
    \multirow{2}{*}{Top 20}
    & \tool & 51.2 &  40.3 & 54.6 & 60.6 & \text{40.6} & \text{48.5} & 63.0  &65.2 & 41.7 & 30.2 & 29.9 & 43.1  \\
    & \oursm & \textbf{80.5} & \textbf{59.6} & \textbf{74.0} & \textbf{75.5} & \textbf{67.1}  & \textbf{63.9} & \textbf{74.5} & \textbf{71.7} & \textbf{47.9} & \textbf{55.6}  & \textbf{37.9} & \textbf{48.9} \\
    \midrule
    \multirow{2}{*}{Top 100} 
    & \tool &32.8 &27.1 & 47.2& 47.4& 30.8 & \text{45.1} & 46.5 & 54.9 & 37.7 &  12.6 & 19.6 & 11.8 \\
    & \oursm & \textbf{71.2}&\textbf{53.4} & \textbf{70.0} & \textbf{67.7} & \textbf{58.9}& \textbf{59.1} & \textbf{67.3}  & \textbf{68.0} & \textbf{45.7} & \textbf{46.2} & \textbf{34.3} & \textbf{45.5} \\
    \bottomrule
    \end{tabular}
\caption
     {Overall performance of \oursm(model trained with \ours) and \tool(models trained under each type of supervision separately) in KILT benchmark. Datasets with * on top indicate in-domain datasets. Top 20 and Top 100 show the performance when given a corpus of size 20 and 100, respectively.} 
\label{table: kilt.main}
\end{table*}

\begin{table}[h]
\centering
\fontsize{6.5}{10} \selectfont
    \begin{tabular}{cc|cc|cc}
    \toprule
    & & \multicolumn{2}{c}{ASQA*} & \multicolumn{2}{c}{ELI5} \\ 
    \midrule
    & & Cor & Gr & Cor & Gr \\
    \midrule
    \multirow{2}{*}{Top 5}
    & \tool  & 28.4 & 42.5 & \textbf{9.7} & 8.9 \\
    & \oursm & \textbf{31.8} & \textbf{44.0} & 9.3 & \textbf{10.3} \\
    \midrule
    \multirow{2}{*}{Top 100}
    &\tool & 20.2 & 31.1 & 9.9 & 13.1 \\
    &\oursm & \textbf{26.3} & \textbf{36.8}  & \textbf{10.5} & \textbf{13.4} \\
    \bottomrule
    \end{tabular}
\caption
     {Overall performance of \oursm and \tool in ALCE benchmarks. Dataset with * on top indicates in-domain datasets. Top 5 and Top 100 show the performance when given a corpus of size 5 and 100, respectively. As the benchmark does not provide gold passages, we skip the retrieval performance.} 
\label{table: alce.main}
\end{table}

Table~\ref{table: kilt.main} and Table~\ref{table: alce.main} show the overall performance on the KILT and ALCE benchmark, respectively.\footnote{ELI5 in the ALCE benchmark is a reformulated version from the KILT benchmark where it contains a smaller number of instances and the evaluation metric is different. For more details on reformulation, please refer to \citet{gao2023enabling}}. Both tables show that models trained with \ours consistently outperform models trained under each type of supervision separately. This suggests that this co-supervision encourages a broader generalization capability throughout the model. In this section, we delve into the impact of each type of supervision on the model's performance and explore the effects of co-supervision. \textbf{From now, for simplicity we name the model trained with \ours as \oursm and the model trained under each supervision separately as \tool\footnote{Please note that \tool is not only trained with NTP but also NSP but separately. We name it NTP for simplicity.}.} More details of each paragraphs are in Appendix~\ref{app: results}.

\paragraph{Co-supervision makes \ctx more robust}
Training a language model with both supervisions from \ntp and \nsp consistently shows higher retrieval performance (average of +16.6) over those trained only with \nsp. Especially, the performance gap tends to increase as corpus size increases and over out-of-domain (improvement rate of 35.04 for in-domain and 41.67 for out-of-domain). Such results suggest that the co-supervision makes \ctx, a language model that constructs the nonparametric sequence space through its output embeddings, more robust. We could further see that it shows more robust performance over other \ctx which is trained via NSP with more datasets and larger batch size (Appendix~\ref{app: nsp}).
As previous research has shown that new embeddings (tokens) can easily adapt to well-established parametric token embedding space of the model~\citep{Hao2023ToolkenGPTAF, Schick2023ToolformerLM}, we hypothesize such adaptability applies to nonparametric sequence embedding space; the robustness of the parametric token embedding space, established during the pretraining step provides a solid foundation that enhances the stability of the nonparametric space. 

\paragraph{Co-supervision enhances generation generalization}

Models trained with \ours consistently outperform those trained under single supervision in terms of correctness and grounding performance across both KILT and ALCE benchmarks. Such results suggest that co-supervision enhances generation generalization of \gen. The performance gap is particularly noticeable in datasets such as FEVER, T-REx, and zsRE. Such results suggest that models under co-supervision demonstrate higher generalization ability showing a broader understanding across diverse input distributions, whereas models trained solely with next token prediction (NTP) struggle with unique input formats. For example, T-REx and zsRE are slot-filling tasks, which present a unique input format ``subject [SEP] relationship type'' which is less likely to be encountered during standard training phases. Additionally, in FEVER, the issue of low grounding performance arises from a poor understanding of instruction. Despite being instructed to generate evidence-based responses, the model tends to generate simple answers without grounding to external knowledge.

\begin{figure}[t!]
    \centering
    \begin{minipage}[b]{0.45\textwidth}
    \includegraphics[width=\textwidth]{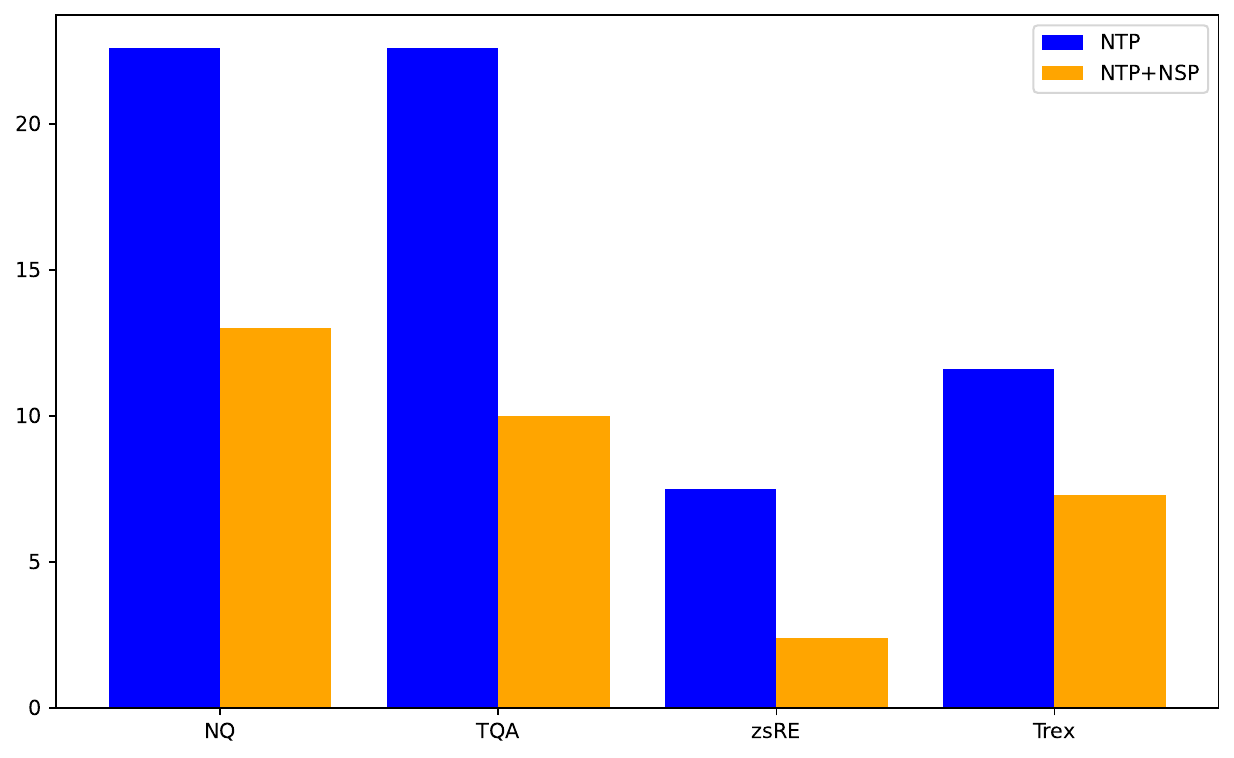}
    \end{minipage}
\caption{Reduction rate of correctness when considering those correct by parametric knowledge as wrong.} 
\label{fig:correctness}
\end{figure}

\paragraph{Co-supervision encourages high interaction between the two spaces}
\label{result: gen}

Figure~\ref{fig:correctness} (Numbers are in Table~\ref{table: correctness} of Appendix) shows the degradation rate of correctness when removing the ones that are correct by the parametric knowledge (categorizing responses associated with incorrect paragraphs as wrong). 
Models trained with \ours show less degradation compared to the one with separate supervision, highlighting that co-supervising encourages interaction between token and sequence embedding space as it is trained to share a common space.
Also, it suggests that the model leverages the contextual knowledge within the nonparametric space for generating responses, as supported by previous research~\citep{min2022nonparametric, lee2023nonparametric}, rather than depending exclusively on its parametric knowledge base.
Even with the restriction of \gen to the parametric space during inference, thereby excluding access to the nonparametric sequence embedding space, the model trained with co-supervision experiences a substantial decline in correctness (from 45.7 to 32.8) in contrast to the model solely supervised with NTP (from 32.8 to 32.0). This suggests that \ours promotes the utilization of knowledge from the nonparametric sequence space rather than relying solely on memorization, effectively leveraging information from its parametric space.

\paragraph{Flowing loss over the sequences when calculating \ntp in co-supervision tends to make the model memorize the knowledge rather than utilizing the knowledge from nonparametric embedding space}
When we do not mask sequences when calculating \ntp of token-sequence co-supervision, \gen tend to rely more on their memorized parametric knowledge whereas models trained with the sequences masked tend to utilize the external knowledge retrieved from the nonparametric sequence embedding space. 
Such a tendency aligns with the findings of \citet{Mallen2022WhenNT}, where the model tends to depend on retrieved knowledge more when it lacks familiarity with the information (long-tail knowledge). By calculating \ntp over the sequences, the model tends to encode the context knowledge in its parameters, leading to a reduced reliance on retrieved knowledge and more on its own knowledge. 

\paragraph{How does the choice of \ctx affect performance?}
\begin{figure}[ht!]
    \centering
    \begin{minipage}[b]{0.45\textwidth}
    \includegraphics[width=\textwidth]{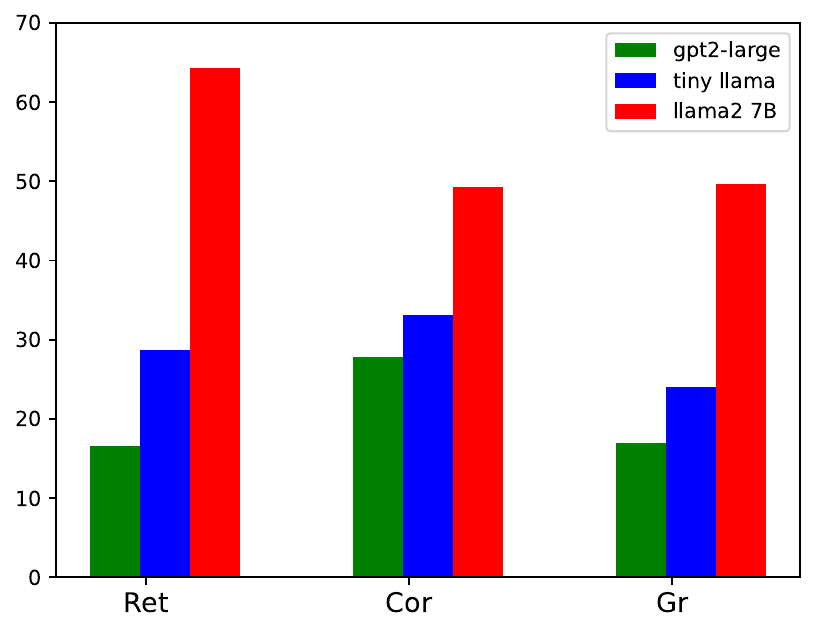}
    \end{minipage}
\caption{Overall performance of how different \ctx, which constructs the nonparametric sequence embedding space, affects the overall performance when training with \oursm. We experiment over 3 different models, GPT2-large, TinyLlama, Llama2-7B.} 
\label{fig:diff_ctx}
\end{figure}

We investigate how using different pretrained models for \ctx impacts overall performance as the nonparametric sequence embedding space is constructed through the output embeddings of \ctx. We compare results over three different pretrained models for \ctx: GPT2-large~\citep{Radford2019gpt2}, TinyLlama~\citep{zhang2024tinyllama}, and Llama2-7B. As shown in Figure~\ref{fig:diff_ctx}, Llama2-7B shows the highest performance. The result suggests that the specific distribution inherent to each pretrained language model influences the performance. As \ours trains a model in a multi-task manner thereby constructing a common space of parametric token space and nonparametric sequence space through co-supervision, the training process appears to be most stable when \gen and \ctx are derived from the same model, thus sharing the same distribution. This observation underlines the significance of distribution compatibility between \gen and \ctx in enhancing model training and performance.

\paragraph{Affect of weight lambda ($\lambda$) when training \ours}
\begin{figure}[t!]
    \centering
    \begin{minipage}[b]{0.4\textwidth}
    \includegraphics[width=\textwidth]{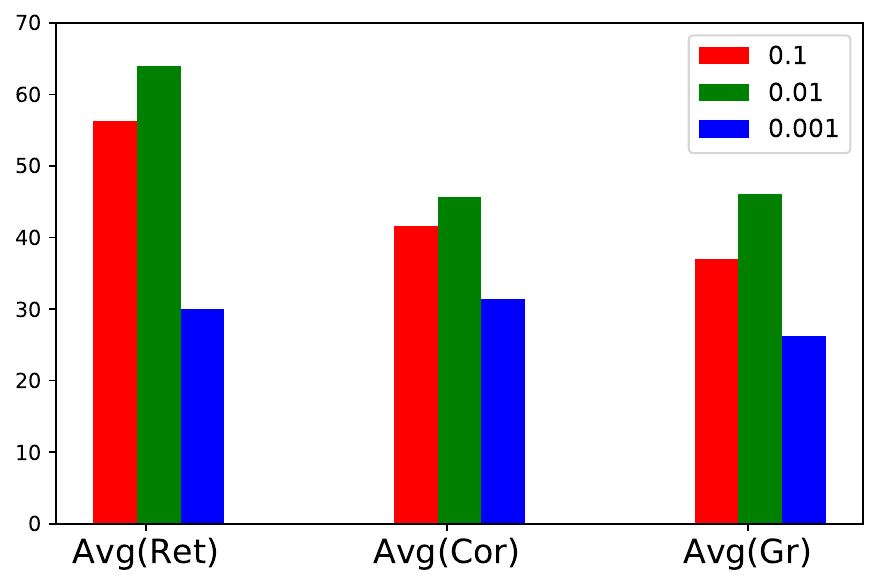}
    \end{minipage}
\caption{Average performance of each metric over 8 datasets in KILT when changing weight parameter $\lambda$ of \oursm.} 
\label{fig:NP_weight}
\end{figure}

We investigate how different weights (\{$10^{-1}$, $10^{-2}$, $10^{-3}$\}) between the next token prediction supervision (\ntp) and next sequence prediction supervision (\nsp) affect the model's performance, which is the lambda weight in Equation~\ref{eq: cosup_loss}. In our setup, as a single generation model \gen receives co-supervision from both the parametric token embedding space and the nonparametric sequence embedding space, the weight determines the balance of influence between these two spaces on the model's training. 
Figure~\ref{fig:NP_weight} (numbers in Table~\ref{table: np_weight} in the Appendix) illustrates that a weight of $10^{-3}$ results in poor retrieval performance, indicating challenges in grasping and stabilizing the nonparametric sequence embedding space. Moreover, at a weight of $10^{-1}$, the model's generation ability tends to decline, suggesting that it rather ruins the well-formed parametric token embedding space. This analysis underscores the critical role of balancing supervision from both embedding spaces to optimize model performance across various metrics.

\section{Conclusion}
In this paper, we propose a semiparametric token-sequence co-supervision training method, which trains a single autoregressive language model with both supervision from parametric token embedding space and nonparametric sequence embedding space in a simultaneous manner. Experiments over 10 information-seeking datasets show that such co-supervision consistently outperforms models trained with each supervision separately of average +14.2, demonstrating that constructing a common space through co-supervision fosters the generalization and robustness of the language model. Such a method is not only limited to sequence space but can be expandable to any embedding space which we leave as future work.

\section{Limitation}
Due to resource constraints, our experimentation did not extend to altering \gen with other pretrained LMs such as Mistral~\citep{jiang2023mistral}.
While we have identified several indicators suggesting that training \gen and \ctx through \ours enhances robustness compared to training each model separately, further exploration with increased computational resources would provide a more comprehensive understanding.

\bibliography{anthology,custom}

\begin{thebibliography}{48}
\expandafter\ifx\csname natexlab\endcsname\relax\def\natexlab#1{#1}\fi

\bibitem[{Alayrac et~al.(2022)Alayrac, Donahue, Luc, Miech, Barr, Hasson, Lenc, Mensch, Millican, Reynolds et~al.}]{alayrac2022flamingo}
Jean-Baptiste Alayrac, Jeff Donahue, Pauline Luc, Antoine Miech, Iain Barr, Yana Hasson, Karel Lenc, Arthur Mensch, Katherine Millican, Malcolm Reynolds, et~al. 2022.
\newblock Flamingo: a visual language model for few-shot learning.
\newblock \emph{Advances in Neural Information Processing Systems}, 35:23716--23736.

\bibitem[{Asai et~al.(2023)Asai, Wu, Wang, Sil, and Hajishirzi}]{asai2023selfrag}
Akari Asai, Zeqiu Wu, Yizhong Wang, Avirup Sil, and Hannaneh Hajishirzi. 2023.
\newblock \href {https://arxiv.org/abs/2310.11511} {{Self-RAG}: Learning to retrieve, generate, and critique through self-reflection}.
\newblock \emph{arXiv preprint arXiv:2310.11511}.

\bibitem[{Bansal et~al.(2024)Bansal, Samanta, Dalmia, Gupta, Vashishth, Ganapathy, Bapna, Jain, and Talukdar}]{bansal2024llm}
Rachit Bansal, Bidisha Samanta, Siddharth Dalmia, Nitish Gupta, Shikhar Vashishth, Sriram Ganapathy, Abhishek Bapna, Prateek Jain, and Partha Talukdar. 2024.
\newblock Llm augmented llms: Expanding capabilities through composition.
\newblock \emph{arXiv preprint arXiv:2401.02412}.

\bibitem[{Borgeaud et~al.(2022)Borgeaud, Mensch, Hoffmann, Cai, Rutherford, Millican, Van Den~Driessche, Lespiau, Damoc, Clark et~al.}]{borgeaud2022improving}
Sebastian Borgeaud, Arthur Mensch, Jordan Hoffmann, Trevor Cai, Eliza Rutherford, Katie Millican, George~Bm Van Den~Driessche, Jean-Baptiste Lespiau, Bogdan Damoc, Aidan Clark, et~al. 2022.
\newblock Improving language models by retrieving from trillions of tokens.
\newblock In \emph{International conference on machine learning}, pages 2206--2240. PMLR.

\bibitem[{Brown et~al.(2020)Brown, Mann, Ryder, Subbiah, Kaplan, Dhariwal, Neelakantan, Shyam, Sastry, Askell et~al.}]{brown2020language}
Tom Brown, Benjamin Mann, Nick Ryder, Melanie Subbiah, Jared~D Kaplan, Prafulla Dhariwal, Arvind Neelakantan, Pranav Shyam, Girish Sastry, Amanda Askell, et~al. 2020.
\newblock Language models are few-shot learners.
\newblock \emph{Advances in neural information processing systems}, 33:1877--1901.

\bibitem[{Devlin et~al.(2018)Devlin, Chang, Lee, and Toutanova}]{devlin2018bert}
Jacob Devlin, Ming-Wei Chang, Kenton Lee, and Kristina Toutanova. 2018.
\newblock Bert: Pre-training of deep bidirectional transformers for language understanding.
\newblock \emph{arXiv preprint arXiv:1810.04805}.

\bibitem[{Dinan et~al.(2019)Dinan, Roller, Shuster, Fan, Auli, and Weston}]{Dinan2018WizardOW}
Emily Dinan, Stephen Roller, Kurt Shuster, Angela Fan, Michael Auli, and Jason Weston. 2019.
\newblock Wizard of wikipedia: Knowledge-powered conversational agents.
\newblock In \emph{ICLR}.

\bibitem[{ElSahar et~al.(2018)ElSahar, Vougiouklis, Remaci, Gravier, Hare, Laforest, and Simperl}]{ElSahar2018TRExAL}
Hady ElSahar, Pavlos Vougiouklis, Arslen Remaci, Christophe Gravier, Jonathon~S. Hare, Fr{\'e}d{\'e}rique Laforest, and Elena Paslaru~Bontas Simperl. 2018.
\newblock T-rex: A large scale alignment of natural language with knowledge base triples.
\newblock In \emph{LREC}.

\bibitem[{Fan et~al.(2019)Fan, Jernite, Perez, Grangier, Weston, and Auli}]{Fan2019ELI5LF}
Angela Fan, Yacine Jernite, Ethan Perez, David Grangier, Jason Weston, and Michael Auli. 2019.
\newblock Eli5: Long form question answering.
\newblock \emph{ArXiv}.

\bibitem[{Gao et~al.(2023)Gao, Yen, Yu, and Chen}]{gao2023enabling}
Tianyu Gao, Howard Yen, Jiatong Yu, and Danqi Chen. 2023.
\newblock Enabling large language models to generate text with citations.

\bibitem[{Guu et~al.(2020)Guu, Lee, Tung, Pasupat, and Chang}]{guu2020retrieval}
Kelvin Guu, Kenton Lee, Zora Tung, Panupong Pasupat, and Mingwei Chang. 2020.
\newblock Retrieval augmented language model pre-training.
\newblock In \emph{International conference on machine learning}, pages 3929--3938. PMLR.

\bibitem[{Hao et~al.(2023)Hao, Liu, Wang, and Hu}]{Hao2023ToolkenGPTAF}
Shibo Hao, Tianyang Liu, Zhen Wang, and Zhiting Hu. 2023.
\newblock \href {https://api.semanticscholar.org/CorpusID:258823133} {Toolkengpt: Augmenting frozen language models with massive tools via tool embeddings}.
\newblock \emph{ArXiv}, abs/2305.11554.

\bibitem[{Honovich et~al.(2022)Honovich, Aharoni, Herzig, Taitelbaum, Kukliansy, Cohen, Scialom, Szpektor, Hassidim, and Matias}]{honovich2022true}
Or~Honovich, Roee Aharoni, Jonathan Herzig, Hagai Taitelbaum, Doron Kukliansy, Vered Cohen, Thomas Scialom, Idan Szpektor, Avinatan Hassidim, and Yossi Matias. 2022.
\newblock True: Re-evaluating factual consistency evaluation.
\newblock \emph{arXiv preprint arXiv:2204.04991}.

\bibitem[{Izacard et~al.(2021)Izacard, Caron, Hosseini, Riedel, Bojanowski, Joulin, and Grave}]{izacard2021contriever}
Gautier Izacard, Mathilde Caron, Lucas Hosseini, Sebastian Riedel, Piotr Bojanowski, Armand Joulin, and Edouard Grave. 2021.
\newblock \href {https://doi.org/10.48550/ARXIV.2112.09118} {Unsupervised dense information retrieval with contrastive learning}.

\bibitem[{Jiang et~al.(2023)Jiang, Sablayrolles, Mensch, Bamford, Chaplot, Casas, Bressand, Lengyel, Lample, Saulnier et~al.}]{jiang2023mistral}
Albert~Q Jiang, Alexandre Sablayrolles, Arthur Mensch, Chris Bamford, Devendra~Singh Chaplot, Diego de~las Casas, Florian Bressand, Gianna Lengyel, Guillaume Lample, Lucile Saulnier, et~al. 2023.
\newblock Mistral 7b.
\newblock \emph{arXiv preprint arXiv:2310.06825}.

\bibitem[{Joshi et~al.(2020)Joshi, Chen, Liu, Weld, Zettlemoyer, and Levy}]{joshi2020spanbert}
Mandar Joshi, Danqi Chen, Yinhan Liu, Daniel~S Weld, Luke Zettlemoyer, and Omer Levy. 2020.
\newblock Spanbert: Improving pre-training by representing and predicting spans.
\newblock \emph{Transactions of the association for computational linguistics}, 8:64--77.

\bibitem[{Joshi et~al.(2017)Joshi, Choi, Weld, and Zettlemoyer}]{Joshi2017TriviaQAAL}
Mandar Joshi, Eunsol Choi, Daniel~S. Weld, and Luke Zettlemoyer. 2017.
\newblock Triviaqa: A large scale distantly supervised challenge dataset for reading comprehension.
\newblock In \emph{ACL}.

\bibitem[{Karpukhin et~al.(2020)Karpukhin, O{\u{g}}uz, Min, Lewis, Wu, Edunov, Chen, and Yih}]{karpukhin2020dense}
Vladimir Karpukhin, Barlas O{\u{g}}uz, Sewon Min, Patrick Lewis, Ledell Wu, Sergey Edunov, Danqi Chen, and Wen-tau Yih. 2020.
\newblock Dense passage retrieval for open-domain question answering.
\newblock \emph{arXiv preprint arXiv:2004.04906}.

\bibitem[{Keskar et~al.(2017)Keskar, Mudigere, Nocedal, Smelyanskiy, and Tang}]{keskar2017on}
Nitish~Shirish Keskar, Dheevatsa Mudigere, Jorge Nocedal, Mikhail Smelyanskiy, and Ping Tak~Peter Tang. 2017.
\newblock \href {https://openreview.net/forum?id=H1oyRlYgg} {On large-batch training for deep learning: Generalization gap and sharp minima}.
\newblock In \emph{International Conference on Learning Representations}.

\bibitem[{Khandelwal et~al.(2019)Khandelwal, Levy, Jurafsky, Zettlemoyer, and Lewis}]{khandelwal2019generalization}
Urvashi Khandelwal, Omer Levy, Dan Jurafsky, Luke Zettlemoyer, and Mike Lewis. 2019.
\newblock Generalization through memorization: Nearest neighbor language models.
\newblock \emph{arXiv preprint arXiv:1911.00172}.

\bibitem[{Kwiatkowski et~al.(2019)Kwiatkowski, Palomaki, Redfield, Collins, Parikh, Alberti, Epstein, Polosukhin, Devlin, Lee, Toutanova, Jones, Kelcey, Chang, Dai, Uszkoreit, Le, and Petrov}]{Kwiatkowski2019NaturalQA}
Tom Kwiatkowski, Jennimaria Palomaki, Olivia Redfield, Michael Collins, Ankur~P. Parikh, Chris Alberti, Danielle Epstein, Illia Polosukhin, Jacob Devlin, Kenton Lee, Kristina Toutanova, Llion Jones, Matthew Kelcey, Ming-Wei Chang, Andrew~M. Dai, Jakob Uszkoreit, Quoc~V. Le, and Slav Petrov. 2019.
\newblock Natural questions: A benchmark for question answering research.
\newblock \emph{TACL}.

\bibitem[{Lee et~al.(2023{\natexlab{a}})Lee, Joo, Kim, Jang, Kim, On, and Seo}]{Lee2023HowWD}
Hyunji Lee, Sejune Joo, Chaeeun Kim, Joel Jang, Doyoung Kim, Kyoung-Woon On, and Minjoon Seo. 2023{\natexlab{a}}.
\newblock \href {https://api.semanticscholar.org/CorpusID:265212695} {How well do large language models truly ground?}
\newblock \emph{ArXiv}, abs/2311.09069.

\bibitem[{Lee et~al.(2023{\natexlab{b}})Lee, Kim, Chang, Oh, Yang, Karpukhin, Lu, and Seo}]{lee2023nonparametric}
Hyunji Lee, Jaeyoung Kim, Hoyeon Chang, Hanseok Oh, Sohee Yang, Vladimir Karpukhin, Yi~Lu, and Minjoon Seo. 2023{\natexlab{b}}.
\newblock Nonparametric decoding for generative retrieval.
\newblock In \emph{Findings of the Association for Computational Linguistics: ACL 2023}, pages 12642--12661.

\bibitem[{Lester et~al.(2021)Lester, Al-Rfou, and Constant}]{lester2021power}
Brian Lester, Rami Al-Rfou, and Noah Constant. 2021.
\newblock The power of scale for parameter-efficient prompt tuning.
\newblock \emph{arXiv preprint arXiv:2104.08691}.

\bibitem[{Levy et~al.(2017)Levy, Seo, Choi, and Zettlemoyer}]{Levy2017ZeroShotRE}
Omer Levy, Minjoon Seo, Eunsol Choi, and Luke Zettlemoyer. 2017.
\newblock Zero-shot relation extraction via reading comprehension.
\newblock In \emph{CoNLL}.

\bibitem[{Lewis et~al.(2020)Lewis, Perez, Piktus, Petroni, Karpukhin, Goyal, Kuttler, Lewis, tau Yih, Rockt{\"a}schel, Riedel, and Kiela}]{Lewis2020RetrievalAugmentedGF}
Patrick Lewis, Ethan Perez, Aleksandara Piktus, Fabio Petroni, Vladimir Karpukhin, Naman Goyal, Heinrich Kuttler, Mike Lewis, Wen tau Yih, Tim Rockt{\"a}schel, Sebastian Riedel, and Douwe Kiela. 2020.
\newblock \href {https://api.semanticscholar.org/CorpusID:218869575} {Retrieval-augmented generation for knowledge-intensive nlp tasks}.
\newblock \emph{ArXiv}, abs/2005.11401.

\bibitem[{Li and Liang(2021)}]{li2021prefix}
Xiang~Lisa Li and Percy Liang. 2021.
\newblock Prefix-tuning: Optimizing continuous prompts for generation.
\newblock \emph{arXiv preprint arXiv:2101.00190}.

\bibitem[{Lin et~al.(2023)Lin, Chen, Chen, Shi, Lomeli, James, Rodriguez, Kahn, Szilvasy, Lewis, Zettlemoyer, and Yih}]{Lin2023RADITRD}
Xi~Victoria Lin, Xilun Chen, Mingda Chen, Weijia Shi, Maria Lomeli, Rich James, Pedro Rodriguez, Jacob Kahn, Gergely Szilvasy, Mike Lewis, Luke Zettlemoyer, and Scott Yih. 2023.
\newblock \href {https://api.semanticscholar.org/CorpusID:263605962} {Ra-dit: Retrieval-augmented dual instruction tuning}.
\newblock \emph{ArXiv}, abs/2310.01352.

\bibitem[{Liu et~al.(2023)Liu, Li, Wu, and Lee}]{liu2023llava}
Haotian Liu, Chunyuan Li, Qingyang Wu, and Yong~Jae Lee. 2023.
\newblock Visual instruction tuning.

\bibitem[{Loshchilov and Hutter(2019)}]{loshchilov2019decoupled}
Ilya Loshchilov and Frank Hutter. 2019.
\newblock \href {http://arxiv.org/abs/1711.05101} {Decoupled weight decay regularization}.

\bibitem[{Mallen et~al.(2022)Mallen, Asai, Zhong, Das, Hajishirzi, and Khashabi}]{Mallen2022WhenNT}
Alex Mallen, Akari Asai, Victor Zhong, Rajarshi Das, Hannaneh Hajishirzi, and Daniel Khashabi. 2022.
\newblock \href {https://api.semanticscholar.org/CorpusID:260443047} {When not to trust language models: Investigating effectiveness and limitations of parametric and non-parametric memories}.
\newblock \emph{ArXiv}, abs/2212.10511.

\bibitem[{Min et~al.(2022)Min, Shi, Lewis, Chen, Yih, Hajishirzi, and Zettlemoyer}]{min2022nonparametric}
Sewon Min, Weijia Shi, Mike Lewis, Xilun Chen, Wen-tau Yih, Hannaneh Hajishirzi, and Luke Zettlemoyer. 2022.
\newblock Nonparametric masked language modeling.
\newblock \emph{arXiv preprint arXiv:2212.01349}.

\bibitem[{Pappas et~al.(2020)Pappas, Mulcaire, and Smith}]{pappas2020grounded}
Nikolaos Pappas, Phoebe Mulcaire, and Noah~A Smith. 2020.
\newblock Grounded compositional outputs for adaptive language modeling.
\newblock \emph{arXiv preprint arXiv:2009.11523}.

\bibitem[{Petroni et~al.(2021)Petroni, Piktus, Fan, Lewis, Yazdani, De~Cao, Thorne, Jernite, Karpukhin, Maillard, Plachouras, Rockt{\"a}schel, and Riedel}]{petroni-etal-2021-kilt}
Fabio Petroni, Aleksandra Piktus, Angela Fan, Patrick Lewis, Majid Yazdani, Nicola De~Cao, James Thorne, Yacine Jernite, Vladimir Karpukhin, Jean Maillard, Vassilis Plachouras, Tim Rockt{\"a}schel, and Sebastian Riedel. 2021.
\newblock \href {https://doi.org/10.18653/v1/2021.naacl-main.200} {{KILT}: a benchmark for knowledge intensive language tasks}.
\newblock In \emph{Proceedings of the 2021 Conference of the North American Chapter of the Association for Computational Linguistics: Human Language Technologies}, pages 2523--2544, Online. Association for Computational Linguistics.

\bibitem[{Radford et~al.(2019)Radford, Wu, Child, Luan, Amodei, and Sutskever}]{Radford2019gpt2}
Alec Radford, Jeff Wu, Rewon Child, David Luan, Dario Amodei, and Ilya Sutskever. 2019.
\newblock \href {https://api.semanticscholar.org/CorpusID:160025533} {Language models are unsupervised multitask learners}.

\bibitem[{Schick et~al.(2023)Schick, Dwivedi-Yu, Dess{\`i}, Raileanu, Lomeli, Zettlemoyer, Cancedda, and Scialom}]{Schick2023ToolformerLM}
Timo Schick, Jane Dwivedi-Yu, Roberto Dess{\`i}, Roberta Raileanu, Maria Lomeli, Luke Zettlemoyer, Nicola Cancedda, and Thomas Scialom. 2023.
\newblock \href {https://api.semanticscholar.org/CorpusID:256697342} {Toolformer: Language models can teach themselves to use tools}.
\newblock \emph{ArXiv}, abs/2302.04761.

\bibitem[{Shi et~al.(2023)Shi, Min, Yasunaga, Seo, James, Lewis, Zettlemoyer, and Yih}]{shi2023replug}
Weijia Shi, Sewon Min, Michihiro Yasunaga, Minjoon Seo, Rich James, Mike Lewis, Luke Zettlemoyer, and Wen-tau Yih. 2023.
\newblock Replug: Retrieval-augmented black-box language models.
\newblock \emph{arXiv preprint arXiv:2301.12652}.

\bibitem[{Thorne et~al.(2018)Thorne, Vlachos, Christodoulopoulos, and Mittal}]{Thorne2018FEVERAL}
James Thorne, Andreas Vlachos, Christos Christodoulopoulos, and Arpit Mittal. 2018.
\newblock Fever: a large-scale dataset for fact extraction and verification.
\newblock In \emph{NACCL}.

\bibitem[{Touvron et~al.(2023{\natexlab{a}})Touvron, Martin, Stone, Albert, Almahairi, Babaei, Bashlykov, Batra, Bhargava, Bhosale et~al.}]{touvron2023llama}
Hugo Touvron, Louis Martin, Kevin Stone, Peter Albert, Amjad Almahairi, Yasmine Babaei, Nikolay Bashlykov, Soumya Batra, Prajjwal Bhargava, Shruti Bhosale, et~al. 2023{\natexlab{a}}.
\newblock Llama 2: Open foundation and fine-tuned chat models.
\newblock \emph{arXiv preprint arXiv:2307.09288}.

\bibitem[{Touvron et~al.(2023{\natexlab{b}})Touvron, Martin, Stone, Albert, Almahairi, Babaei, Bashlykov, Batra, Bhargava, Bhosale, Bikel, Blecher, Ferrer, Chen, Cucurull, Esiobu, Fernandes, Fu, Fu, Fuller, Gao, Goswami, Goyal, Hartshorn, Hosseini, Hou, Inan, Kardas, Kerkez, Khabsa, Kloumann, Korenev, Koura, Lachaux, Lavril, Lee, Liskovich, Lu, Mao, Martinet, Mihaylov, Mishra, Molybog, Nie, Poulton, Reizenstein, Rungta, Saladi, Schelten, Silva, Smith, Subramanian, Tan, Tang, Taylor, Williams, Kuan, Xu, Yan, Zarov, Zhang, Fan, Kambadur, Narang, Rodriguez, Stojnic, Edunov, and Scialom}]{Touvron2023Llama2O}
Hugo Touvron, Louis Martin, Kevin~R. Stone, Peter Albert, Amjad Almahairi, Yasmine Babaei, Nikolay Bashlykov, Soumya Batra, Prajjwal Bhargava, Shruti Bhosale, Daniel~M. Bikel, Lukas Blecher, Cristian~Cant{\'o}n Ferrer, Moya Chen, Guillem Cucurull, David Esiobu, Jude Fernandes, Jeremy Fu, Wenyin Fu, Brian Fuller, Cynthia Gao, Vedanuj Goswami, Naman Goyal, Anthony~S. Hartshorn, Saghar Hosseini, Rui Hou, Hakan Inan, Marcin Kardas, Viktor Kerkez, Madian Khabsa, Isabel~M. Kloumann, A.~V. Korenev, Punit~Singh Koura, Marie-Anne Lachaux, Thibaut Lavril, Jenya Lee, Diana Liskovich, Yinghai Lu, Yuning Mao, Xavier Martinet, Todor Mihaylov, Pushkar Mishra, Igor Molybog, Yixin Nie, Andrew Poulton, Jeremy Reizenstein, Rashi Rungta, Kalyan Saladi, Alan Schelten, Ruan Silva, Eric~Michael Smith, R.~Subramanian, Xia Tan, Binh Tang, Ross Taylor, Adina Williams, Jian~Xiang Kuan, Puxin Xu, Zhengxu Yan, Iliyan Zarov, Yuchen Zhang, Angela Fan, Melanie Kambadur, Sharan Narang, Aurelien Rodriguez, Robert Stojnic, Sergey Edunov, and
  Thomas Scialom. 2023{\natexlab{b}}.
\newblock \href {https://api.semanticscholar.org/CorpusID:259950998} {Llama 2: Open foundation and fine-tuned chat models}.
\newblock \emph{ArXiv}, abs/2307.09288.

\bibitem[{Wolf et~al.(2019)Wolf, Debut, Sanh, Chaumond, Delangue, Moi, Cistac, Rault, Louf, Funtowicz, and Brew}]{Wolf2019HuggingFacesTS}
Thomas Wolf, Lysandre Debut, Victor Sanh, Julien Chaumond, Clement Delangue, Anthony Moi, Pierric Cistac, Tim Rault, R{\'e}mi Louf, Morgan Funtowicz, and Jamie Brew. 2019.
\newblock \href {https://api.semanticscholar.org/CorpusID:208117506} {Huggingface's transformers: State-of-the-art natural language processing}.
\newblock \emph{ArXiv}, abs/1910.03771.

\bibitem[{Yang et~al.(2017)Yang, Dai, Salakhutdinov, and Cohen}]{yang2017breaking}
Zhilin Yang, Zihang Dai, Ruslan Salakhutdinov, and William~W Cohen. 2017.
\newblock Breaking the softmax bottleneck: A high-rank rnn language model.
\newblock \emph{arXiv preprint arXiv:1711.03953}.

\bibitem[{Yang et~al.(2018)Yang, Qi, Zhang, Bengio, Cohen, Salakhutdinov, and Manning}]{Yang2018HotpotQAAD}
Zhilin Yang, Peng Qi, Saizheng Zhang, Yoshua Bengio, William~W. Cohen, Ruslan Salakhutdinov, and Christopher~D. Manning. 2018.
\newblock Hotpotqa: A dataset for diverse, explainable multi-hop question answering.
\newblock In \emph{EMNLP}.

\bibitem[{Yoon et~al.(2024)Yoon, Jang, Kim, Kim, Shafayat, and Seo}]{yoon2024langbridge}
Dongkeun Yoon, Joel Jang, Sungdong Kim, Seungone Kim, Sheikh Shafayat, and Minjoon Seo. 2024.
\newblock Langbridge: Multilingual reasoning without multilingual supervision.
\newblock \emph{arXiv preprint arXiv:2401.10695}.

\bibitem[{Zhang et~al.(2024)Zhang, Zeng, Wang, and Lu}]{zhang2024tinyllama}
Peiyuan Zhang, Guangtao Zeng, Tianduo Wang, and Wei Lu. 2024.
\newblock Tinyllama: An open-source small language model.
\newblock \emph{arXiv preprint arXiv:2401.02385}.

\bibitem[{Zhang et~al.(2022)Zhang, Roller, Goyal, Artetxe, Chen, Chen, Dewan, Diab, Li, Lin et~al.}]{zhang2022opt}
Susan Zhang, Stephen Roller, Naman Goyal, Mikel Artetxe, Moya Chen, Shuohui Chen, Christopher Dewan, Mona Diab, Xian Li, Xi~Victoria Lin, et~al. 2022.
\newblock Opt: Open pre-trained transformer language models.
\newblock \emph{arXiv preprint arXiv:2205.01068}.

\bibitem[{Zhao et~al.(2023)Zhao, Gu, Varma, Luo, Huang, Xu, Wright, Shojanazeri, Ott, Shleifer, Desmaison, Balioglu, Damania, Nguyen, Chauhan, Hao, Mathews, and Li}]{zhao2023fsdp}
Yanli Zhao, Andrew Gu, Rohan Varma, Liang Luo, Chien-Chin Huang, Min Xu, Less Wright, Hamid Shojanazeri, Myle Ott, Sam Shleifer, Alban Desmaison, Can Balioglu, Pritam Damania, Bernard Nguyen, Geeta Chauhan, Yuchen Hao, Ajit Mathews, and Shen Li. 2023.
\newblock \href {http://arxiv.org/abs/2304.11277} {Pytorch fsdp: Experiences on scaling fully sharded data parallel}.

\bibitem[{Zhong et~al.(2022)Zhong, Lei, and Chen}]{zhong2022training}
Zexuan Zhong, Tao Lei, and Danqi Chen. 2022.
\newblock Training language models with memory augmentation.
\newblock \emph{arXiv preprint arXiv:2205.12674}.

\end{thebibliography}
\bibliographystyle{acl_natbib}

\appendix
\section{Implementation Details} \label{app: implementation_details}
\subsection{Gathering in-batch negatives} \label{app: train}
Suppose we have $B$ batched instances $D_1, \ldots D_B$ for each training step, and each $D_i$ has $c_{i,1}, \ldots c_{i, l_i}$ reference context($l_i \geq 1$). We can collect $M = \sum_{i=1}^{B}l_i$ context embeddings(after counting duplicates) and the same amount of query embedding, each referring to one positive target. We set the target context embedding as the positive one, while setting the rest $M-1$ as the in-batch negative samples.

During the experiments, we gather all the context embeddings across all 8 GPUs to increase the number of in-batch negatives. We ensure at least 63 \footnote{at least 1 context per instance, 8 per each GPU, a total of 8 GPUs infer 64 total reference context embeddings, yielding at least $64-1=63$ in-batch negatives.}, but the exact number differs across training steps (experimentally about 80 to 90).

\subsection{Generation by Grounding in Inference Step} \label{app: generation_by_grounding}

To distinguish where the model generates by grounding on the retrieved sequence and where the model generates in a freeform, we add two special tokens \cs and \ce which each indicate the start of the generation by grounding on the retrieved sequence and the end, respectively. In other words, when the model generates a response in the form of ``$q, n_1, \ret, c_1, g_1, \cdots n_i, \ret, c_i, g_i, \cdots $'', the part between \cs and \ce is the $g_i$ part and the rest is $n_i$ indicating freeform generation. 

\section{Experiments Setup}
\label{sec:exp_setup}
\subsection{Details of metrics}
We assess the generation results in three axes: correctness, retrieval performance, and grounding performance
\paragraph{Correctness}
Correctness evaluates how well the model answers the given query for each task. For each dataset, we chose the metric to evaluate following the metric used in its official paper. Details for each dataset is in Table~\ref{table: correctness_metric}.
\paragraph{Retrieval Performance}
Retrieval performance measures whether the model retrieves relevant paragraphs to answer the given question. We measure in two aspects, whether the gold paragraph exists within retrieved paragraphs (Ret) and retrieval precision to assess how many of the model-retrieved paragraphs contain gold paragraphs (Ret-P). For example, when the model retrieves three different paragraphs \{P1, P2, P3\} while generating a response and only one of them \{P2\} is the gold paragraph, Ret will be 100 since there is a gold paragraph in the set of retrieved paragraphs while Ret-P is $\frac{1}{3}$ since only one is correct. Please note that we measure the metric by considering both gold and retrieved as a set; when the same paragraph is retrieved twice, we consider it as one during the calculation.
\paragraph{Grounding Performance}
Grounding performance evaluates how well the model generates based on given external knowledge. Following the approach of previous works~\citep{gao2023enabling, Lee2023HowWD}, we use TRUE~\citep{honovich2022true}, a T5-11B model finetuned on various NLI datasets, to see whether the external knowledge entails a part of the response that is generated based on the knowledge. 
To be more specific, as we add special tokens \cs and \ce to distinguish between natural form generation and generation based on given external knowledge (grounding generation); \cs indicates the starting point of grounding generation, and \ce indicates the endpoint. (Appendix~\ref{app: generation_by_grounding})
Thereby we parse the generated response into pairs (external knowledge, grounding generation) and calculate the average of whether the external knowledge entails the grounding generation in sentence wise of grounding generation. 

\subsection{Details of training dataset}

\begin{table}[t!]
\centering
\fontsize{7.5}{10}\selectfont
    \begin{tabular}{ccc}
    \toprule
    Source & Name  & Instance Num \\
    \midrule
    \multirow{5}{*}{Open-Instruct} 
    & GPT-4 Alpaca & 6363\\
    & Stanford Alpaca & 7826\\
    & FLAN-V2 & 720\\
    & ShareGPT & 2544\\
    & Open Assistant 1 & 2671\\
    \midrule
    \multirow{2}{*}{KILT} 
    & Wizard of Wikipedia & 2159\\
    & Natural Questions & 12306\\
    & FEVER & 5502\\
    \midrule
    \multirow{2}{*}{HF Dataset} 
    & OpenBookQA &54\\
    & Arc-Easy & 183\\
    \midrule
    ASQA & ASQA & 2732\\ 
    \bottomrule
    \end{tabular}
\caption
     {\fontsize{6.5}{10}\footnotesize Statistic and characteristics of training dataset (42932 instances). Instance Num inside the parentheses indicate the number of corresponding instances within all the filtered instances (43060 instances)}
\label{table: train_dataset}
\end{table}

\paragraph{Filtering}
We apply 3 filtering from \citet{asai2023selfrag} to make our training dataset.
\begin{enumerate}
    \item Relevance and Support: We retained instances with relevant and fully or partially supported citations to ensure dataset integrity.
    \item Citation Count: Instances without citations or with more than 3 citations were filtered out to balance complexity and informativeness.
    \item Length Constraint: We removed instances exceeding 1024 tokens (input + output) to optimize computational efficiency.

\end{enumerate}
These criteria yielded a final training dataset of 42932 instances, poised to provide high-quality examples for effective model training.

\paragraph{Statistics}
Among 42.9k training instances, 32.1k instances contain 1 \ret, 6.7k instances contain 2 \ret, and 4k instances contain 3 \ret in the output.
The source of these instances is shown in Table~\ref{table: train_dataset}.

\paragraph{Single \ret Example}
Question: "which river has the largest basin in world"

Answer:
"\ret Amazon River and Crenarchaeota. The Amazon has over 1,100 tributaries, 12 of which are over long. Some of the more notable ones are: Amazon River The Amazon River (, ; Spanish and ) in South America is the largest river by discharge volume of water in the world, and by some definitions it is the longest. The headwaters of the Apurímac River on Nevado Mismi had been considered for nearly a century as the Amazon's most distant source, until a 2014 study found it to be the Cordillera Rumi Cruz at the headwaters of the Mantaro River in Peru. The Mantaro and Question: which river has the largest basin in world \cs Amazon \ce"

\paragraph{Multi \ret Example}
Question: "Why do corporations lobby for laws that impoverish lower class citizens?"

Answer: 
"\ret Corporate capitalism power and influence corporations and large business interest groups have over government policy, including the policies of regulatory agencies and influencing political campaigns (see corporate welfare). Many social scientists have criticized corporations for failing to act in the interests of the people, and their existence seems to circumvent the principles of democracy, which assumes equal power relations between individuals in a society. In an April 29, 1938 message to the Congress, Franklin D. Roosevelt warned that the growth of private power could lead to fascism: Dwight D. Eisenhower criticized the notion of the confluence of corporate power and \"de facto\" Question: Why do corporations lobby for laws that impoverish lower class citizens? \cs While it is true that corporations may lobby for laws or regulations that are favorable to their own interests, it is not necessarily the case that these laws or regulations would have negative impacts on lower class citizens. \ce \ret Anti-corporate activism control the markets, that income inequality is increasing, and that corporations have gained too much power. As part of the political left, activists against corporate power and influence support a decreased income gap and improved economical equity. The defenders of corporations such as Ron Arnold highlight that governments do legislate in ways that restrict the actions of corporations (see Sarbanes-Oxley Act) and that lawbreaking companies and executives are routinely caught and punished, usually in the form of monetary fines. In addition, from the perspective of business ethics it might be argued that chief executives are not inherently more evil than Question: Why do corporations lobby for laws that impoverish lower class citizens? \cs For example, a corporation may lobby for laws that reduce regulation on its industry, which could potentially lead to lower costs and higher profits for the corporation, but could also have negative consequences for workers or consumers. \ce It is important to recognize that the relationship between corporations, lobbying, and public policy is complex, and it is not always clear how specific laws or regulations will impact different groups of people.In general, it is important for citizens to stay informed about the activities of corporations and to advocate for policies that benefit the common good.",

\subsection{Details of evaluation dataset}

\paragraph{Construction Step}
Following the evaluation setup of \citet{gao2023enabling}, we retrieve the top 100 paragraphs from the Wikipedia corpus provided by KILT for each instance in the dataset. We retrieve paragraphs by utilizing a well-performing retriever model, \texttt{contriever-msmarco}~\citep{izacard2021contriever}. 
When constructing the top 100 paragraphs, we initially populate the corpus set with gold annotations from the KILT benchmark and subsequently supplement the remainder with paragraphs retrieved from \texttt{contriever-msmarco}, ensuring that all gold paragraphs are in the top 100 paragraphs.
\paragraph{Dataset Statistics and Characteristics}
In Table~\ref{table: eval_dataset}, we present the statistics and characteristics of the datasets in KILT~\citep{petroni-etal-2021-kilt} benchmark employed for evaluation. Datasets lacking evidence annotation, such as WNED-CWEB, WNED-WIKI, and AIDA CoNLL-YAGO, are excluded from the KILT benchmark.

\subsection{Training Details}
The default experiment setting for both \ours and the baselines is set to train the pre-trained Llama2 7B~\citep{Touvron2023Llama2O} provided from huggingface~\citep{Wolf2019HuggingFacesTS} as the initial model for both \gen and \ctx. We use 8 Nvidia A100 with 80GB memory for our experiments. We set the maximum token length to be 1,024 due to the memory constraint. We use FSDP~\citep{zhao2023fsdp} to conduct multi-GPU distributed training. We set the base hyperparameter as epoch 3, batch size 8, and AdamW optimizer~\citep{loshchilov2019decoupled} with no decay, and learning rate of 2e-5 and a decayed rate gamma of 0.85 every 1 epoch. For \ours, we set the weight $\lambda$ as 0.01 while training. We also apply gradient clipping to the \gen model, with a maximum norm equal to 1.
We run inference of our trained models using 1 A6000 GPU with 48GB memory.

\subsection{Instructions for evaluation}
Table~\ref{table: instruction} shows the instructions used during evaluation. For those datasets where it is used in \citet{asai2023selfrag} or \citet{gao2023enabling}, we follow or reformulate the instruction so that the model provides the evidence to measure the grounding performance.
\begin{table*}[t!]
\centering
\fontsize{7.5}{9}\selectfont
\caption{\fontsize{7.5}{10}\footnotesize Instructions used during evaluation.} 
\begin{tabular}{ m{2cm} m{13cm}} 
    \toprule
    \textbf{Dataset} & \textbf{Instruction}  \\
    \midrule
        WoW & Given a chat history separated by new lines, generates an informative, knowledgeable and engaging response. \\
        FEVER & Is the following statement correct or not? Say supports if it's correct; otherwise say refutes. Also provide the evidence. \\
        ELI5 & Provide a paragraph-length response using simple words to answer the following question. \\
        NQ, TriviaQA  & Answer the following question with corresponding evidence. \\
        zsRE, T-REx & Find the correct entity given subject and relation with corresponding evidence. \\
        HotpotQA & Answer question that require 2 step reasoning where each reasoning steps have corresponding evidence. \\
        ASQA & Answer the following question. The question may be ambiguous and have multiple correct answers, and in that case, you have to provide a long-form answer including all correct answers. \\
    \bottomrule
\end{tabular}
\label{table: instruction}
\end{table*}

\begin{table*}[t!]
\centering
\fontsize{7.5}{10}\selectfont
    \begin{tabular}{c|ccc|ccccccc}
    \toprule
    Name & Task & Instance Num & Input Format & Output Format & Reference\\
    \midrule
    Natural Questions (NQ) & ODQA  & 2,837 & Question & Extractive & \citet{Kwiatkowski2019NaturalQA} \\
    Wizard of Wikipedia (WoW) & Dialogue Conversation  & 3,054 & Long& Abstractive & \citet{Dinan2018WizardOW} \\
    FEVER & Fact Checking & 10,444 & Claim & Classification & \citet{Thorne2018FEVERAL} \\
    \midrule
    TriviaQA & ODQA & 5,359 & Question & Extractive & \citet{Joshi2017TriviaQAAL} \\
    ELI5 &  ODQA & 1,507 &  Question & Long Abstractive & \citet{Fan2019ELI5LF} \\
    Zero Shot RE (zsRE) & Slot Filling  & 3,724 & Structured & Entity & \citet{Levy2017ZeroShotRE} \\
    T-REx & Slot Filling  & 5,000 & Structured & Entity & \citet{ElSahar2018TRExAL} \\
    HotpotQA & ODQA  & 5,600 &  Question & Short Abstractive & \citet{Yang2018HotpotQAAD} \\
    \bottomrule
    \end{tabular}
\caption
     {\fontsize{6.5}{10}\footnotesize Statistic and characteristics of evaluation dataset.}
\label{table: eval_dataset}
\end{table*}

\begin{table}[t!]
\centering
\fontsize{7.5}{10}\selectfont
    \begin{tabular}{c|ccc|ccccccc}
    \toprule
    Name & Metric \\
    \midrule
    Natural Questions (NQ) & Answer Accuracy \\
    Wizard of Wikipedia (WoW) & Unigram F1 \\
    FEVER & NLI \\
    \midrule
    TriviaQA & Answer Accuracy  \\
    ELI5 & Rougel \\
    Zero Shot RE (zsRE) & Answer Accuracy \\
    T-REx & Answer Accuracy  \\
    HotpotQA & Answer Accuracy \\
    \bottomrule
    \end{tabular}
\caption
     {\fontsize{6.5}{10}\footnotesize Correctness metric for each datasets}
\label{table: correctness_metric}
\end{table}

\section{Experimental Results} \label{app: results}

\begin{table*}[t!]
\centering
\fontsize{6.5}{10} \selectfont
    \begin{tabular}{cc|cc|cc|cc|cc}
    \toprule
     & & Cor. & Cor$^{-}$ &  Cor. & Cor$^{-}$ & Cor. & Cor$^{-}$ & Cor. & Cor$^{-}$ \\
    \midrule
    & & \multicolumn{2}{c}{NQ*} &  \multicolumn{2}{c}{zsRE} & \multicolumn{2}{c}{T-REx} & \multicolumn{2}{c}{TriviaQA} 
     \\
    \midrule
    \multirow{2}{*}{Top20} 
    & \tool & 49.7 & 40.5 & 40.3 & 37.5 & 40.6 & 36.4 & 65.2 & 53.1   \\ 
    & \oursm & 55.7 &49.3 & 59.6 &  58.1 &67.1 & 62.5 & 71.7 & 65.2 \\
    \midrule
    \multirow{2}{*}{Top100} 
    & \tool & 38.1 & 27.7 &  27.1 & 24.3 & 30.8 & 25.8 & 54.9 & 39.2 \\ 
    & \oursm &  50.5 & 44.4 &  53.4 & 52.0 & 58.9&54.6 & 68.0 & 59.8\\
    \bottomrule
    \end{tabular}
\caption
     {Performance of correctness performance (Cor.) and correctness when considering those instances where retrieval fail as incorrect ($\text{Cor}^{-}$)} 
\label{table: correctness}
\end{table*}

\begin{table*}[h]
\centering
\fontsize{6.5}{10} \selectfont
    \begin{tabular}{cc|ccc|ccc|ccc|ccc}
    \toprule
     & & Ret & Cor & Gr &  Ret & Cor & Gr & Ret & Cor & Gr & Ret & Cor & Gr\\
    \midrule
    Force \ret & & \multicolumn{3}{c}{NQ*} & \multicolumn{3}{c}{WoW*} & \multicolumn{3}{c}{FEVER*} & \multicolumn{3}{c}{zsRE}   \\
    \midrule
    \multirow{2}{*}{X} 
    & \tool & 99.9 & 73.7 & 67.4 & 100.0 & 19.4 & 57.1 & 100.0 & 59.1 & 6.0 & 99.9 & 72.4 & 74.7  \\
    & \oursm &  99.8 & 68.6 & 69.3 & 100.0 & 19.6 & 69.8 & 100.0 & 65.4 & 31.0 & 100.0 & 71.7 & 80.8\\
    \midrule
    \multirow{2}{*}{O} 
    & \tool & 100.0 & 71.9 & 69.5 & 100.0 & 19.3 & 59.4 & 100.0 & 59.1 & 6.0 & 100.0 & 70.3 & 74.8 \\ 
    & \oursm &100.0 & 70.9 & 70.7 & 100.0 & 19.9 & 69.5 & 100.0 & 65.6 & 31.7 & 100.0 & 71.8 & 80.3 \\   
    \midrule
    \midrule
     & & \multicolumn{3}{c}{T-REx} & \multicolumn{3}{c}{TriviaQA}  & \multicolumn{3}{c}{ELI5}& \multicolumn{3}{c}{\emph{Avg}}\\
    \midrule
    \multirow{2}{*}{X} 
    & \tool &98.0 & 58.0 & 56.0 & 96.7 & 73.2 & 48.9 & 100.0 & 20.7 & 6.5& 99.2 & 53.8 & 45.2\\
    & \oursm & 100.0 & 78.2 & 71.8 & 100.0 & 68.0 & 45.4 & 100.0 & 20.7 & 6.8 &100.0 & \textbf{56.0} & \textbf{53.5} \\
    \midrule
    \multirow{2}{*}{O} 
    &\tool & 100.0 & 66.2 & 71.0 & 100.0 & 72.4 & 48.6 & 100.0 & 20.7 & 6.5& 100.0 &54.3 & 47.9\\ 
    &\oursm & 100.0 & 78.9 & 72.0 & 100.0 & 68.4 & 44.2 & 100.0 & 20.7 & 6.3 & 100.0 & \textbf{56.6} & \textbf{53.5} \\   
    \bottomrule
    \end{tabular}
\caption
     {Performance over the oracle setup. We skip HotpotQA as the dataset requires two paragraphs, making it difficult to make in the same setting} 
\label{table: oracle_full}
\end{table*}

\begin{table*}[t!]
\centering
\fontsize{6.5}{10} \selectfont
    \begin{tabular}{cc|ccc|ccc|ccc|ccc|ccc}
    \toprule
     & & Ret & Cor & Gr &  Ret & Cor & Gr & Ret & Cor & Gr & Ret & Cor & Gr & Ret & Cor & Gr\\
    \midrule
    Force \ret & & \multicolumn{3}{c}{NQ*} & \multicolumn{3}{c}{zsRE} & \multicolumn{3}{c}{T-REx} & \multicolumn{3}{c}{TriviaQA}  & \multicolumn{3}{c}{\emph{Avg}}   \\
    \midrule
    \multirow{2}{*}{X} 
    & \tool & 0.0 & 12.2 & 13.8 & 0.0 & 7.9 & 15.7 & 0.0 & 9.3 & 7.9 & 0.0 & 7.6 & 8.3 & 0.0& \textbf{9.2}&\textbf{11.4}\\
    & \oursm & 0.0 & 11.7 & 13.6 & 0.0 & 7.3 & 11.6 & 0.0 & 9.3 & 6.6 & 0.0 & 7.6 & 5.2 &0.0	&9.0&	9.2\\
    \midrule
    \multirow{2}{*}{O} 
    & \tool & 0.0 & 11.9 & 15.6 & 0.0 & 7.6 & 16.8 & 0.0 & 8.8 & 8.1 & 0.0 & 7.7 & 8.7&0.0&	\textbf{9.0}&\textbf{12.3} \\ 
    & \oursm & 0.0 & 11.9 & 13.6  & 0.0 & 7.3 & 11.7 & 0.0 & 8.9 & 7.1 & 0.0 & 7.4 & 5.0& 0.0	&\text{8.9}&9.3 \\     
    \bottomrule
    \end{tabular}
\caption
     {Experiment over the case where retrieval always fail. We skip HotpotQA as the dataset requires two paragraphs, making it difficult to make in the same setting, and datasets without answers as it is hard to distinguish false negatives.} 
\label{table: fail}
\end{table*}

\begin{table*}[t!]
\centering
\fontsize{6.5}{10} \selectfont
    \begin{tabular}{c|ccccc|ccccc|ccccc|ccccc}
    \toprule
      BS &  R & R-P & Cor & C$^{-}$ & Gr & R & R-P& Cor& C$^{-}$  &Gr & R& R-P & Cor& C$^{-}$  & Gr  & R &R-P& Cor& C$^{-}$  & Gr \\
    \midrule
    \multicolumn{21}{l}{\textbf{Top20}} \\
    \midrule
     & \multicolumn{5}{c}{NQ*} & \multicolumn{5}{c}{WoW*} & \multicolumn{5}{c}{FEVER*} & \multicolumn{5}{c}{ELI5} \\
    \midrule
     2 &  63.1 & 59.1 & 50.0 & 44.6 & 60.6 & 37.5 & 36.8 & 14.4 & 7.3 & 33.5 & 71.4 & 67.5 & 56.3 & 40.7 & 6.0 & 38.9 & 27.0 & 21.8 & 8.4 & 8.7\\
    4 & 63.3 & 59.0 & 50.8 & 44.9 & 49.0 & 40.8 & 40.1 & 15.3 & 8.2 & 31.7 & 73.7 & 69.2 & 66.0 & 50.0 & 21.0 & 35.1 & 25.0 & 22.2 & 7.8 & 5.3 \\
    8 & 65.1 & 62.7 & 55.7 & 49.3 & 62.6 & 49.8 & 49.7 & 15.7 & 10.4 & 63.7 & 77.5 & 75.9 & 65.2 & 51.5 & 28.0 & 36.3 & 29.4 & 21.5 & 7.8 & 8.7\\
    \midrule
     & \multicolumn{5}{c}{zsRE} & \multicolumn{5}{c}{T-REx} & \multicolumn{5}{c}{TriviaQA} & \multicolumn{5}{c}{HotpotQA} \\
    \midrule
     2 & 68.8 & 65.9 & 50.8 & 49.4 & 48.1 & 68.0 & 66.5 & 60.0 & 55.0 & 46.5 & 71.4 & 68.9 & 57.8 & 54.6 & 43.0 & 48.9 & 71.0 & 29.2 & 26.4 & 43.1 \\
    4 & 78.0 & 75.2 & 56.4 & 55.1 & 50.5 & 73.0 & 71.3 & 65.0 & 60.1 & 46.5 & 72.7 & 68.9 & 68.5 & 62.7 & 38.6 & 52.2 & 73.6 & 36.8 & 33.9 & 36.8\\
    8 & 80.5 & 80.2 & 59.6 & 58.1 & 74.0  & \text{75.5} &\text{75.4}& \text{67.1}& 62.5 & \text{63.9} & 74.5 & 73.0 & 71.7 & 65.2 & 47.9 & 55.6 & 79.1 & 37.9 & 35.2 & 48.9\\

    \midrule
    \midrule
    \multicolumn{21}{l}{\textbf{Top100}} \\
    \midrule
     & \multicolumn{5}{c}{NQ*} & \multicolumn{5}{c}{WoW*} & \multicolumn{5}{c}{FEVER*} & \multicolumn{5}{c}{ELI5} \\
    \midrule
    2 & 53.3 & 49.2 & 43.2 & 38.1 & 56.9 & 24.3 & 23.9 & 13.8 & 4.8 & 29.8 & 58.6 & 54.2 & 55.8 & 32.9 & 5.3 & 21.4 & 14.2 & 21.7 & 4.6 & 8.5 \\
    4 & 52.9 & 48.2 & 44.6 & 38.0 & 56.9 & 27.9 & 27.4 & 14.6 & 5.8 & 30.6 & 61.2 & 56.8 & 61.7 & 34.9 & 18.8 & 18.9 & 12.6 & 22.2 & 4.1 & 8.9\\
    8 & 56.8 & 53.8 & 50.5 & 44.4 & 58.7 & 36.9 & 36.9 & 14.8 & 7.9 & 61.3 & 66.2 & 64.2 & 64.3 & 43.4 & 26.1 & 21.0 & 15.8 & 21.6 & 4.6 & 10.2\\
    \midrule
     & \multicolumn{5}{c}{zsRE} & \multicolumn{5}{c}{T-REx} & \multicolumn{5}{c}{TriviaQA} & \multicolumn{5}{c}{HotpotQA} \\
    \midrule
    2 & 54.1 & 50.2 & 41.1 & 39.7 & 43.7 & 57.6 & 55.0 & 52.2 & 46.1 & 41.4 & 64.0 & 61.3 & 53.1 & 49.4 & 42.0 & 37.1 & 53.8 & 25.7 & 20.8 & 41.1 \\
    4 &67.0 & 63.3 & 50.3 & 48.6 & 45.9 & 65.3 & 62.8 & 59.2 & 53.7 & 42.0 & 65.2 & 60.9 & 64.4 & 56.9 & 35.6 & 41.2 & 57.8 & 32.6 & 27.6 & 33.3 \\
    8 & 71.2 & 70.7 & 53.4 & 52.0 & 70.0 & 67.7	& 67.5& 58.9& 54.6& 59.1 & 67.3 & 65.6 & 68.0 & 59.8 & 45.7 & 46.2 & 65.3 & 34.3 & 29.7 & 45.5\\
    \bottomrule
    \end{tabular}
\caption
     {Performance by changing batch size (BS).} 
     \vspace{-1em}
\label{table: batch_size}
\end{table*}

\begin{table*}
\centering
\fontsize{7.5}{10}\selectfont
\begin{tblr}{
  width = \linewidth,
  colspec = {Q[150]Q[300]Q[200]},
  row{1} = {c},
  hlines,
}
\textbf{Question}  & \textbf{Retrieved Paragraph} & \textbf{Generated Response} \\
who plays peter in what we do in the shadows & Steve Pemberton :: Steven James Pemberton (born 1 September 1967) is an English actor, comedian and writer, best known as a member of \"The League of Gentlemen\" with Reece Shearsmith, Mark Gatiss, and Jeremy Dyson. Pemberton and Shearsmith also co-wrote and appeared in the sitcom \"Psychoville\" and the comedy-drama \"Inside No. 9\". His other television credits include \"Doctor Who\", \"Benidorm\", \"Blackpool\", \"Shameless\", \"Whitechapel\", \"Happy Valley\" and \"Mapp and Lucia\". & The actor Steve Pemberton plays the role of Peter in the British comedy television show. \\
the basic steps of rational decision making include defining goals and & Rational planning model :: The rational model of decision-making is a process for making sound decisions in policy making in the public sector. Rationality is defined as \u201ca style of behavior that is appropriate to the achievement of given goals, within the limits imposed by given conditions and constraints\u201d. It is important to note the model makes a series of assumptions in order for it to work, such as: & The basic steps of rational decision making include defining goals and then analyzing alternative solutions to determine the best course of action.  \\
\end{tblr}
\caption{Example from NQ}
\label{example: grounding_by_ret}
\end{table*}

\begin{table*}[ht!]
\centering
\fontsize{6.5}{10} \selectfont
    \begin{tabular}{cc|cccc|cccc|cccc|cccc}
    \toprule
    Model & Param Sharing &  Ret & Ret-P & Cor. & C-R & Ret & Ret-P& Cor. & C-R & Ret& Ret-P & Cor. & C-R  & Ret &Ret-P& Cor. & C-R \\
    \midrule
    & & \multicolumn{4}{c}{NQ*} & \multicolumn{4}{c}{WoW*} & \multicolumn{4}{c}{FEVER*} & \multicolumn{4}{c}{ELI5} \\
    \midrule
     Llama2-7B & x  & 53.3 & 50.3 & \textbf{54.2} & 28.0 & \text{28.3} & 28.1  & \textbf{15.2}& 48.0& 61.9 & 60.7 & 58.2 & 5.7 & 18.5 & 14.9 & \textbf{22.1} & 7.3  \\
     Llama2-7B & o  & 35.7 & 34.4 & 38.1 & 43.0 & 14.7 & 14.6 & 13.4 & 40.0 & 31.4 & 30.4 & 56.7 & 5.2 & 12.7 &11.0 & \text{21.9} & 7.0   \\
     contriever-msmarco  & o  & 50.3 &49.8& \text{51.0} & 52.1  & 23.0 &23.0& 14.9 & 42.5  & 39.0 &38.9 & 54.5 & 11.0  & 18.8 & 9.6& \textbf{22.1} & 7.3  \\
     \midrule
     \multicolumn{2}{c}{\oursm}& \textbf{56.8} &\textbf{53.8}& \text{50.5} & \textbf{58.7}  & \textbf{36.9} &\textbf{36.9}& \text{14.8} & \textbf{61.3}  & \textbf{66.2} &\textbf{64.2}& \textbf{64.3} & \textbf{26.1}  &\textbf{21.0} &\textbf{15.8}& 21.6 & \textbf{10.2}  \\
    \midrule
    \midrule
     & & \multicolumn{4}{c}{zsRE} & \multicolumn{4}{c}{T-REx} & \multicolumn{4}{c}{TriviaQA} & \multicolumn{4}{c}{HotpotQA} \\
    \midrule
     Llama2-7B  & x & 49.8 &49.5 &  38.4 & 53.9  & 52.4 &52.4& \text{33.2} & \text{45.7}  &65.1 & 64.1& 66.4&44.8 & 43.1 & 63.3 & 34.0 & 39.3 \\
    Llama2-7B  & o  &32.8&32.7 &27.1 & 47.2& 47.4 & 47.4& 30.8 & \text{45.1} &46.5	&45.8 & 54.9 & 37.7 & 12.6 & 12.4 & 19.6 & 11.8\\
    contriever-msmarco  & o & 45.0 &44.0& 36.2 & 42.2 & 36.9 & 36.7& \text{33.8} & 33.1 & \textbf{67.3} &\text{65.1}&\textbf{73.0} & \text{42.2} & \text{46.0} & 65.1 & 34.0 & \text{37.4}   \\
     \midrule
     \multicolumn{2}{c}{\oursm}& \textbf{71.2} & \textbf{70.7}&\textbf{53.4} & \textbf{70.0} & \textbf{67.7} & \textbf{67.5}& \textbf{58.9}& \textbf{59.1}  & \textbf{67.3} & \textbf{65.6}& 68.0 & \textbf{45.7} & \textbf{46.2}&\textbf{65.3} & \textbf{34.3} & \textbf{45.5} \\
    \bottomrule
    \end{tabular}
\caption
     {Overall Top100 performance of the KILT benchmark when replacing \ctx to other general models trained via NSP. Param Sharing shows whether the \ctx that extracts query embedding and \ctx that extracts context embedding are the same. Llama2-7B models are initialized with Llama2-7B and are further trained with the training dataset whereas contriever-msmarco is the released model from huggingface which is trained via msmarco. } 
\label{table: tool_ret.top100}
\end{table*}

\subsection{Does the benefit of \ours still hold when replacing \ctx trained via NSP to more general retrieval model?} \label{app: nsp}
Table~\ref{table: tool_ret.top100} shows the overall Top100 performance of the KILT benchmark when replacing \ctx to other general models trained via NSP, including the one that is widely known in out-of-domain, contriever-msmarco~\citep{izacard2021contriever}. Llama without parameter sharing tends to show the strongest performance, where we could see that \ctx trained via co-supervision (\oursm) tend to show robust performance over all cases. Such results suggest that the nonparametric sequence embedding space constructed through co-supervision is well-constructed compared to those trained via only NSP.

\subsection{Co-supervision encourages high interaction between the two spaces }
\paragraph{Correctness degradation in cases where retrieval fails}
Table~\ref{table: correctness} shows the degradation of correctness when removing the ones that are correct by the parametric knowledge (categorizing responses associated with incorrect paragraphs as wrong). Models trained with semiparametric token-sequence co-supervision exhibit less degradation compared to those with separate supervision. This suggests that co-supervision promotes interaction between token and sequence embedding space, encouraging the model to share a common space for enhanced performance.

\paragraph{Grounding performance tends to vary by retrieval performance}

When analyzing the impact of retrieval success on grounding performance, \oursm significantly outperforms in grounding when retrieval is successful compared to when it fails, whereas the model trained only by NTP shows little difference regardless of retrieval outcome.
Upon investigating why models trained with \oursm exhibit lower grounding performance upon retrieval failure, it appears to stem from the disconnect caused by attempting to answer the query with the incorrect document. In such cases, the model might fetch information that seems closest to the expected answer from the external knowledge but ends up generating content that also contains knowledge from the given query, leading to less relevance to the retrieved paragraph or fabricating information not present in the paragraph (Examples in Table~\ref{example: grounding_by_ret} in Appendix). 
Conversely, models trained with \tool demonstrate consistent grounding performance, unaffected by the success or failure of retrieval. This suggests that \tool's grounding capability is more reliant on its generative performance rather than the accuracy of retrieval, leading to similar outcomes irrespective of whether the correct information was retrieved or not.
Based on these findings, future work could explore critiquing the success of retrieval based on grounding scores.

\begin{table}[t!]
\centering
\fontsize{6.5}{10} \selectfont
    \begin{tabular}{c|cc|cc}
    \toprule
    & \multicolumn{2}{c}{Oracle} & \multicolumn{2}{c}{Failure} \\ 
    \midrule
    & Cor & Gr & Cor & Gr \\
    \midrule
     \tool  &  54.3 & 47.9 & \textbf{9.0} & \textbf{12.3} \\
     \oursm & \textbf{56.6} & \textbf{53.5} & 8.9 & 9.3 \\
    \bottomrule
    \end{tabular}
\caption
     {Generation performance without the condition of retrieval performance} 
\label{table: oracle}
\end{table}

\paragraph{Generation performance without the condition of retrieval performance}
As correctness and grounding performance depend on retrieval performance, we evaluate both \oursm and \tool trained models in a setting where retrieval is always correct or always wrong to see the correctness and grounding performance without the condition of retrieval performance.
We evaluate two settings where 1. oracle: retrieval is always correct and 2. failure: retrieval is always wrong. 
Table~\ref{table: oracle} shows the average performance of each metric over datasets in the KILT benchmark for oracle setting (all numbers in Table~\ref{table: oracle_full}) and Table~\ref{table: fail} shows performance for failure setting.
\oursm trained models shows higher performance in the oracle setup whereas lower performance in the failure setup.
Such results correlate with the findings on top where since \ours trained tends to get the answer correct based on the retrieved paragraphs, it shows high correctness in the oracle setup whereas lower correctness in the failure setup.
Also as we could see that \ours trained models tend to show low grounding performance when retrieval fails, we can see that it shows lower grounding performance in the failure setup whereas high grounding performance in the oracle setup.

\subsection{Weight parameter $\lambda$}
Table~\ref{table: np_weight} shows the Top20 performance of the KILT dataset with varying values of the weight parameter ($\lambda$) in Equation~\ref{eq: cosup_loss}. Notably, for $\lambda = 10^{-3}$, we could see that the model tends to not converge, resulting in significantly lower performance.

\begin{table*}[t!]
\centering
\fontsize{6.5}{10} \selectfont
    \begin{tabular}{c|cccc|cccc|cccc|cccc}
    \toprule
    Np weight & Ret & Ret-P & Cor. & C-R & Ret & Ret-P& Cor. & C-R & Ret& Ret-P & Cor. & C-R  & Ret &Ret-P& Cor. & C-R \\
    \midrule
    & \multicolumn{4}{c}{NQ*} & \multicolumn{4}{c}{WoW*} & \multicolumn{4}{c}{FEVER*} & \multicolumn{4}{c}{ELI5} \\
    \midrule
    $10^{-1}$ & 59.0 & 55.9 & 47.7 & 43.3 & 42.0 & 41.6 & 15.2 & 36.8 & 67.5 & 64.7 & 64.4 & 36.0 & \textbf{40.7} & 29.0 & 22.5 & 5.2 \\
    \textbf{$10^{-2}$} & \textbf{65.1} & \textbf{62.7} & \textbf{55.7} & \textbf{62.6} & \textbf{49.8} & \textbf{49.7} & \textbf{15.7} & \textbf{63.7} & \textbf{77.5} & \textbf{75.9} & \textbf{65.2} & \textbf{28.0} & \text{36.3} &\textbf{29.4}& 21.5 & \textbf{8.7} \\  
    $10^{-3}$ & 33.7 & 33.6 & 35.7 & 34.9 & 14.0 & 14.0 & 13.2 & 27.0 &  27.4 &27.3 &28.3 & 9.2 & 19.9 & 19.4 & \textbf{22.6}& 4.2\\
    \midrule
    \midrule
     & \multicolumn{4}{c}{zsRE} & \multicolumn{4}{c}{T-REx} & \multicolumn{4}{c}{TriviaQA} & \multicolumn{4}{c}{HotpotQA} \\
    \midrule
    $10^{-1}$ & 64.7 & 64.0 & 39.0 & 47.2 & 64.2 & 63.6 & \text{50.0}& \text{45.2} & 70.2 & 68.0 & 64.9 & 41.1 & 41.7 & 61.0 & 29.2 & 41.8\\
    \textbf{$10^{-2}$} &\textbf{80.5} & \textbf{80.2} & \textbf{59.6} & \textbf{74.0} & \textbf{75.5} &\textbf{75.4}& \textbf{67.1}&\textbf{63.9}& \textbf{74.5} & \textbf{73.0} & \textbf{71.7} & \textbf{47.9} & \textbf{55.6} & \textbf{79.1} & \textbf{37.9} & \textbf{48.9}    \\  
    $10^{-3}$ & 33.6 & 33.6 & 26.2 & 41.8 & 48.5 & 48.4 & 48.0 & 38.3 & 40.3 & 40.2 & 53.9 & 24.9 & 20.7 & 31.1 & 23.9 & 29.9\\
    \bottomrule
    \end{tabular}
\caption
     {Performance by different value of weight parameter $\lambda$.} 
\label{table: np_weight}
\end{table*}

\subsection{Effect of Batch Size}
\begin{figure}[t!]
    \centering
    \begin{minipage}[b]{0.4\textwidth}
    \includegraphics[width=\textwidth]{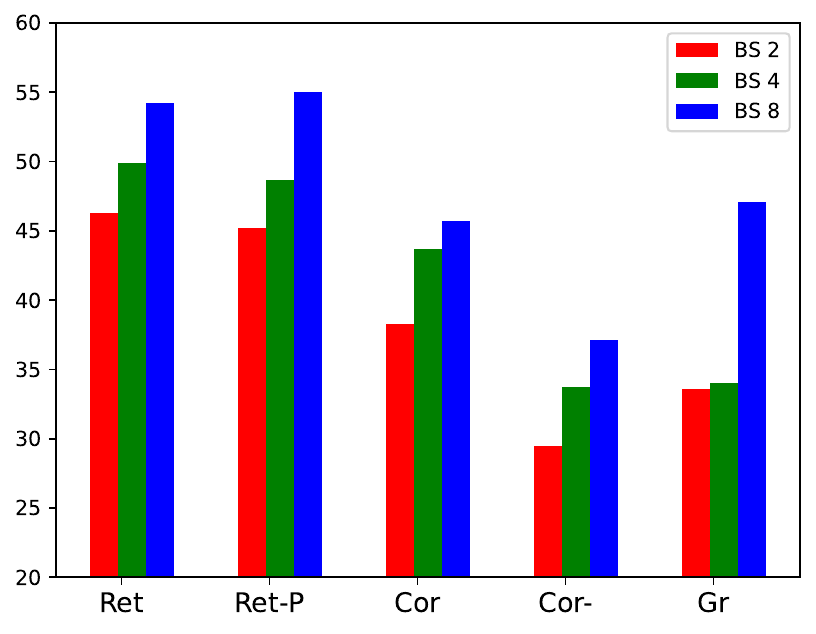}
    \end{minipage}
\caption{\fontsize{6.5}{10}\footnotesize Average performance of each metric over 8 datasets in KILT when changing batch size. Each number indicates batch size per GPU.} 
\label{fig:batch_size}
\end{figure}
Results in Figure~\ref{fig:batch_size} (Numbers in Table~\ref{table: batch_size}) show the average performance of each metric over 8 datasets in KILT with different batch sizes per GPU.
Results show the importance of increasing the batch size in \oursm; performance of all metrics tends to increase with increasing batch size. 
We hypothesize such a trend largely due to stable training of NSP; as it is widely known that retrieval models tend to show higher performance with larger batch size, which is not always true for generation models~\citep{keskar2017on}.
As NSP is calculated via in-batch negatives, training with larger batch size in other words increasing the number of negatives consistently improves retrieval performance~\citep{izacard2021contriever, karpukhin2020dense}. 

\begin{figure}[t!]
    \centering
    \begin{minipage}[b]{0.45\textwidth}
    \includegraphics[width=\textwidth]{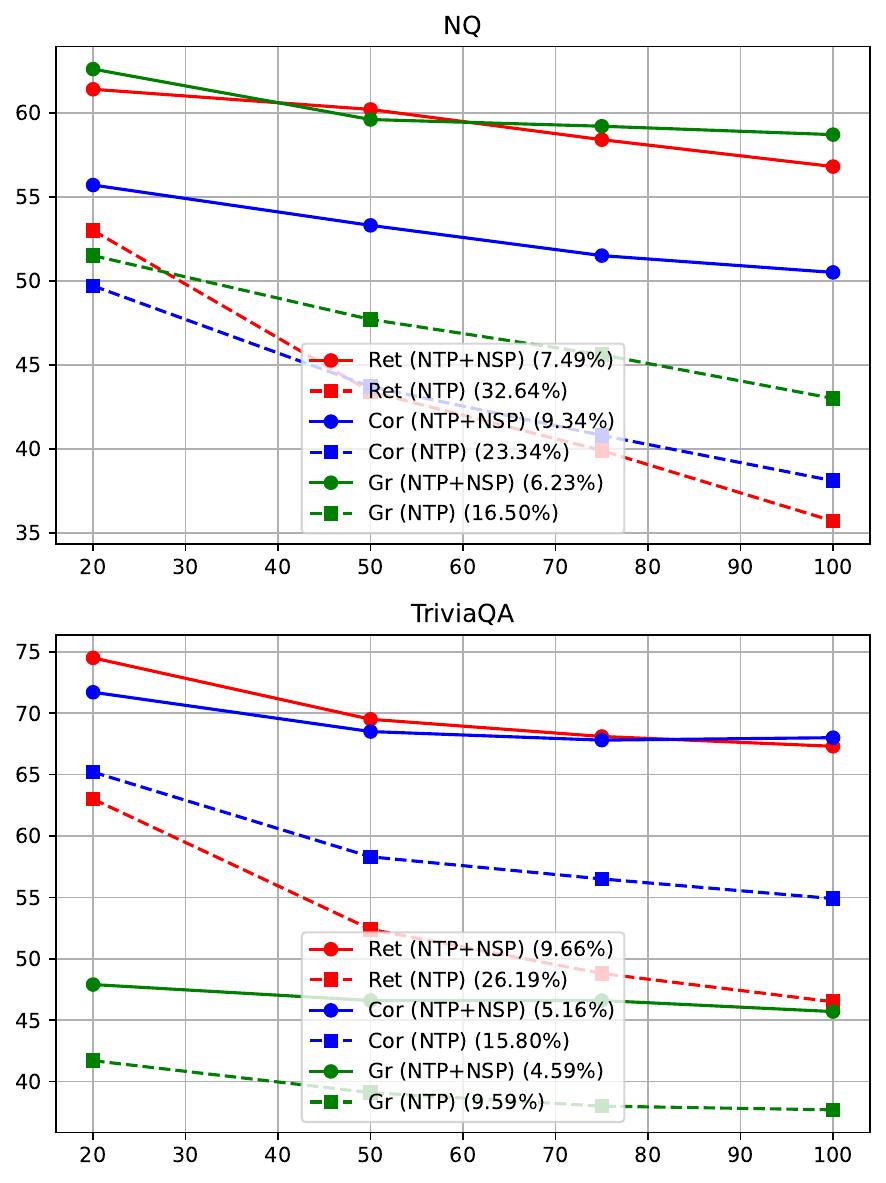}
    \end{minipage}
\caption{\fontsize{6.5}{10}\footnotesize \oursm tend to be more robust with larger corpus size (x-axis)} 
\label{fig:diff_by_ctx_num}
\end{figure}

\subsection{Performance gap tends to increase as corpus size increases} 
Figure~\ref{fig:diff_by_ctx_num} shows that the performance gap between \oursm and \tool tends to increase as corpus size increases; \oursm shows more stable and robust performance with different sizes of the corpus.

\end{document}